\documentclass[11pt]{article}

\usepackage[final]{acl}

\usepackage{times}
\usepackage{latexsym}

\usepackage[T1]{fontenc}

\usepackage[utf8]{inputenc}

\usepackage{microtype}

\usepackage{inconsolata}

\usepackage{graphicx}

\usepackage{float}
\usepackage{amsmath}
\usepackage[dvipsnames]{xcolor}
\usepackage{booktabs}
\usepackage{algorithm}
\usepackage{algpseudocode} 
\usepackage{listings}
\usepackage{xcolor}
\usepackage{graphicx}
\usepackage{subcaption}
\usepackage{placeins} 
\usepackage{multirow}

\DeclareMathOperator*{\argmin}{arg\,min}
\DeclareMathOperator*{\argmax}{arg\,max}

\lstdefinestyle{promptstyle}{
  basicstyle=\ttfamily\footnotesize,
  breaklines=true,
  frame=single,
  columns=fullflexible,
  keepspaces=true,
  showstringspaces=false
}

\title{\raisebox{-0.2\height}{\includegraphics[height=1.2em]{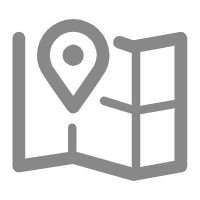}}\,%
MA\textsuperscript{2}P: A Meta-Cognitive Autonomous Intelligent Agents Framework for Complex Persuasion}

\author{
\textbf{Dingyi Zhang\textsuperscript{1}},
\textbf{Ziqing Zhuang\textsuperscript{1}},
\textbf{Linhai Zhang\textsuperscript{2}},
\textbf{Ziyang Gao\textsuperscript{1}},
\textbf{Deyu Zhou\textsuperscript{1}\thanks{Corresponding author: \texttt{d.zhou@seu.edu.cn}.}} \\
\textsuperscript{1}School of Computer Science and Engineering, Key Laboratory of Computer Network \\
and Information Integration, Ministry of Education, Southeast University, China \\
\textsuperscript{2}Department of Informatics, King’s College London \\
\texttt{\{zhangdy, d.zhou\}@seu.edu.cn} \\
}

\begin{document}
\maketitle
\begin{abstract}
Persuasive dialogue generation plays a vital role in decision-making, negotiation, counseling, and behavior change, yet it remains a challenging problem.
In complex persuasion where the persuadee's internal states are not expressed clearly, the persuader must interpret responses, infer the persuadee's latent mental states (e.g., beliefs and desires), and translate them into targeted, strategy-consistent actions; however, current approaches often produce generic or weakly grounded responses even when such cues are identified.
Moreover, although large language models (LLMs) can generate persuasive content, their performance varies substantially across domains due to uneven knowledge coverage and limited reasoning generalization.
To address these challenges, we propose MA\textsuperscript{2}P, a meta-cognitive autonomous intelligent agent framework for complex persuasion. Specifically, we develop an autonomous multi-agent architecture that coordinates perception management, mental-state inference, strategy execution, memory maintenance, and performance evaluation.
To mitigate cross-domain performance variation, we further design a meta-cognitive configurator that selects an appropriate meta-strategy from a structured knowledge base at the outset, thereby guiding subsequent reasoning and planning.
Experimental results show that our approach achieves a higher persuasion success rate than baselines.

\end{abstract}

\section{Introduction}

Persuasive dialogue plays a central role in real-world decision making, including negotiation, healthcare counseling, education, and behavior change \citep{rogiers2024persuasionlargelanguagemodels,SINGH2025100197}. 
Unlike purely informative conversation, persuasion is inherently goal-directed: the persuader must influence another agent's beliefs or actions under uncertainty, which makes the task closer to sequential planning than next-turn language generation \citep{cheng2025strategicpersuasionlanguagemodels,10.1162/TACL.a.16}. 
In this work, we focus on \emph{complex persuasion}, where success depends not only on domain knowledge across diverse topics but also on the ability to model the persuadee's internal state.

Recent research has expanded persuasive dialogue from early single-domain settings \citep{wang-etal-2019-persuasion} to more diverse domains \citep{jin-etal-2024-persuading}, and newer datasets further expose persuadee mental states such as beliefs and desires, raising the bar for personalized, state-aware persuasion \citep{zhang2025persuasiondoubleblindmultidomaindialogue}. Meanwhile, current LLM-based approaches typically rely on a single persuader model to infer user concerns from dialogue and generate the next turn; such models are fluent and can sometimes identify salient constraints or resistance signals \citep{ju2025adaptivepsychologicalpersuasionlarge}. However, as scenarios become more mental-state grounded, the key difficulty shifts from recognizing concerns to acting on them: the persuader must translate inferred states into concrete strategy choices and maintain coherent multi-turn planning, otherwise it often falls back to generic advice or drifts across turns \citep{cheng2025strategicpersuasionlanguagemodels}.

\begin{figure*}[t]
    \centering
    \includegraphics[width=0.95\textwidth]{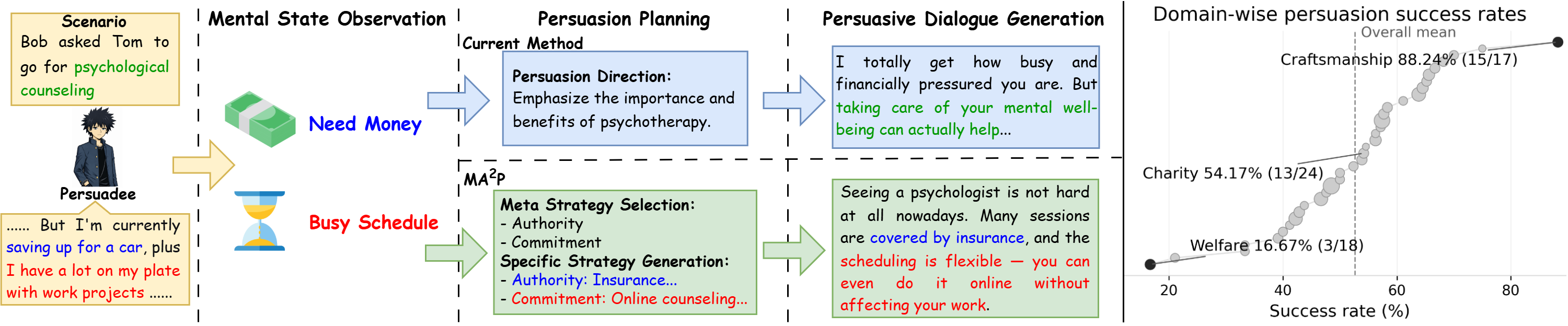}
    \caption{Motivation for MA\textsuperscript{2}P. \textbf{Left:} a CToMPersu example \citep{zhang2025persuasiondoubleblindmultidomaindialogue} where current LLM persuaders identify concerns but fail to respond with strategy-grounded actions. \textbf{Right:} \texttt{gpt-5-mini} success rates on CToMPersu show large cross-domain fluctuations, indicating weak generalization.}
    \label{fig:persuasion}
    \vspace{-0.3cm}
\end{figure*}

Despite recent progress, existing LLM based persuaders still face two key limitations, which motivate our design. 
(1) No actionable planning.
They are often \emph{reactive}: even when they surface concrete obstacles, such as ``need money'' and ``busy schedule'' in Fig.~\ref{fig:persuasion}, they do not convert them into an actionable persuasion plan. 
In the illustrated case, the current approach produces only a high-level direction, ``emphasize the importance and benefits of psychotherapy'', and then generates a generic response, rather than offering targeted solutions aligned with the extracted obstacles, such as insurance coverage and flexible online scheduling. 
(2) Cross-domain fluctuations.
LLM persuaders generalize poorly across domains: on CToMPersu \citep{zhang2025persuasiondoubleblindmultidomaindialogue}, \texttt{gpt-5-mini} success rates range from 88.24\% (best domain) to 16.67\% (worst), a 71.57-point gap in Fig.~\ref{fig:persuasion}. The dispersion is also reflected by a large standard deviation, visually evidenced by the wide horizontal spread of domain points and many domains lying far from the dashed overall-mean line. These motivations call for structured mechanisms that map mental-state cues to strategy-conditioned actions and improve robustness across diverse domains.

To address these challenges, we propose MA\textsuperscript{2}P, a meta-cognitive autonomous-agent framework for persuasive dialogue. Inspired by LeCun et al.'s blueprint for autonomous intelligent agents \citep{LeCun2022APT}, it formulates persuasion as a \emph{closed-loop} interaction and decomposes the persuader into modular components for perception, mental-state inference, strategy planning, response generation, memory, and evaluation, so that detected obstacles can be translated into concrete, strategy-conditioned actions rather than generic next-turn advice. This transfer is non-trivial because persuasion is language- and mental-state-driven, requiring task-specific redesign of the perception, world-model, and actor modules, while the persuadee is an adaptive agent rather than a passive environment. To address these challenges, MA\textsuperscript{2}P combines modularization with a meta-level controller. Inspired by meta-cognition, particularly planning, monitoring, and evaluation \citep{Metcalfe1994MetacognitionK}, MA\textsuperscript{2}P uses a task-specific meta-level mechanism that selects a high-level meta-strategy once at the beginning of each persuasion interaction, guides turn-level strategy generation, and updates experience after the interaction. Across automatic and human evaluations, MA\textsuperscript{2}P consistently improves persuasion success, mental-state alignment, and planning coherence, while reducing cross-domain variance. The prompt design is released.\footnote{\href{https://github.com/DingyiZhang/MAP-A-Meta-Cognitive-Autonomous-Intelligent-Agents-Framework-for-Complex-Persuasion}{MA\textsuperscript{2}P Prompt Design}}

\begin{figure*}[t]
\centering
\includegraphics[width=0.95\textwidth]{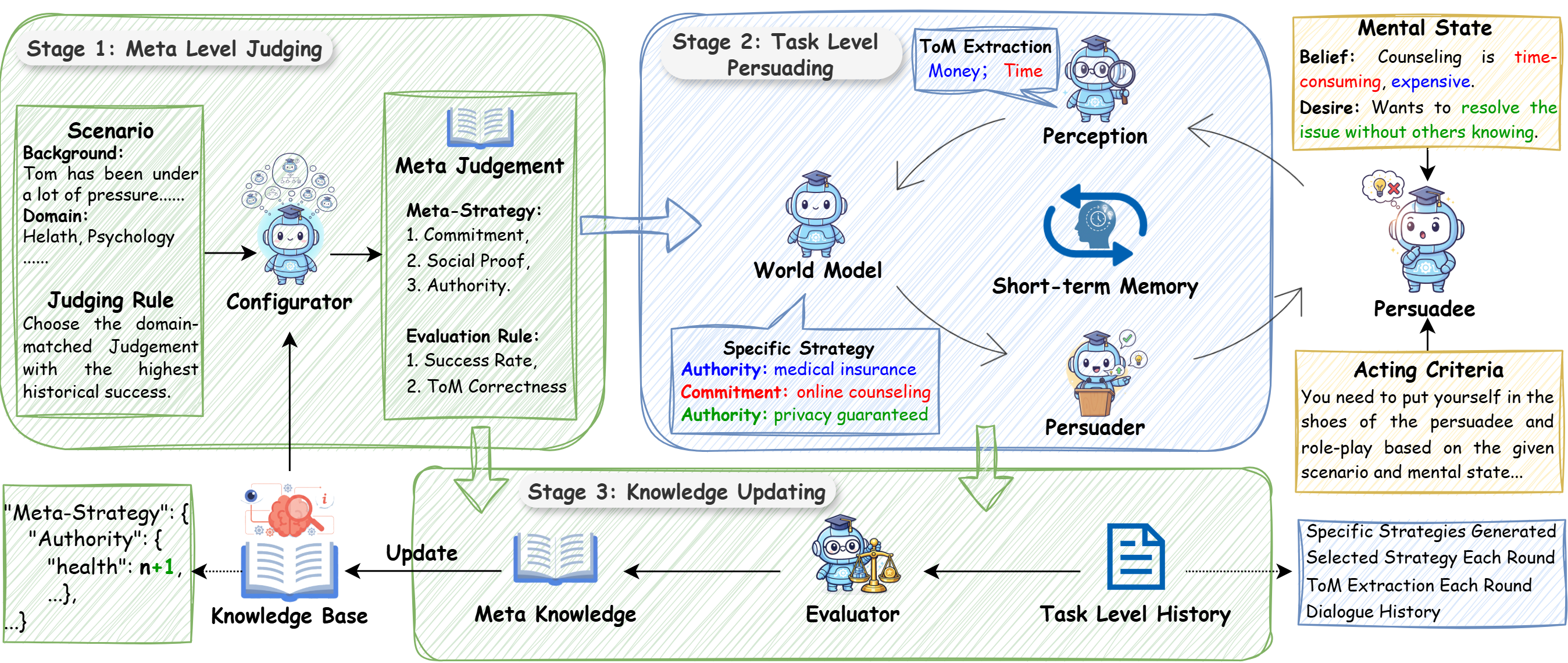} 
\caption{Overview of the proposed \textbf{MA\textsuperscript{2}P} (Meta-Cognitive autonomous intelligent agents) framework for persuasive dialogue. 
It consists of three stages: 
(1) Meta-level Judging, where the Configurator selects a meta-strategy and evaluation rules from the knowledge base; 
(2) Task-level Persuading, where autonomous intelligent agents collaboratively generate persuasion responses; 
(3) Knowledge Updating, where the Evaluator assesses outcomes and successful cases are written back to the knowledge base for future tasks.}
\label{fig:solution_overview}
\vspace{-0.2cm}
\end{figure*}

\section{Background}
\label{sec:background}

\paragraph{Autonomous intelligent agents: purpose and typical use cases.}
Autonomous intelligent agents are designed for \emph{long-horizon, interactive} tasks where decisions must be updated online as new feedback arrives.
They are widely used in settings such as robotics and control, embodied navigation, tool-using assistants, and multi-turn interactive systems, where the agent must maintain state, plan ahead, and correct itself under partial observability \citep{LeCun2022APT}.

\paragraph{A unified definition.}
An autonomous agent can be characterized as a closed-loop decision process that repeatedly (i) \emph{perceives} the environment, (ii) \emph{updates} an internal state, (iii) \emph{selects} an action (optionally via planning), and (iv) \emph{receives feedback} and continues.
Following \citet{LeCun2022APT}, we use $x_t$ for the observation at step $t$, $s_t$ for a latent internal state, and $a_t$ for the action.
A common formulation is:
\[
\begin{aligned}
s_t &= \mathrm{Enc}(x_{\le t}), \\
a_t &\sim \pi(\cdot \mid s_t), \\
s_{t+1} &= \mathrm{Pred}(s_t,a_t),
\end{aligned}
\]
where $\mathrm{Enc}$ summarizes past inputs into a task-relevant state, $\pi$ maps the state to an action, and $\mathrm{Pred}$ (the world model) predicts the next state under an action.
To support goal-directed behavior beyond one-step reactions, the agent is typically guided by a scalar objective (or cost) $C(\cdot)$ that scores states and enables multi-step planning:
\[
\begin{aligned}
\hat{a}_{t:t+T}=\argmin_{a_{t:t+T}} \sum_{\tau=t+1}^{t+T} C(s_\tau)
\quad \text{s.t.}\quad \\
s_{\tau+1}=\mathrm{Pred}(s_\tau,a_\tau).
\end{aligned}
\]

This definition highlights the essential components of autonomy: state estimation ($\mathrm{Enc}$), predictive modeling ($\mathrm{Pred}$), action selection ($\pi$), and objective-driven planning via $C(\cdot)$ \citep{LeCun2022APT}.

\section{Method}
Our framework, MA\textsuperscript{2}P (meta-cognitive autonomous intelligent agents), formulates persuasive dialogue as a three-stage meta-level plan--act--reflect loop. 
As shown in Fig.~\ref{fig:solution_overview}, \textit{Configurator} retrieves a meta-strategy and evaluation rules from the knowledge base (stage 1), a team of autonomous intelligent agents (\textit{Perception}, \textit{World Model}, \textit{Persuader}, \textit{Short-Term Memory}) executes multi-turn persuasion (stage 2), and the \textit{Evaluator} updates the knowledge base with successful patterns (stage 3). 
This update improves subsequent stage 1 decisions by providing stronger domain- and scenario-specific evidence. Algorithm~\ref{alg:ma2p} summarizes the overall procedure.

\subsection{Meta-Level Judging}
Before each persuasion episode, MA\textsuperscript{2}P performs meta-level judging to choose a guiding meta-strategy and specify evaluation rules given the scenario and domain. 
As shown in Fig.~\ref{fig:solution_overview}, stage 1 is carried out by the \textit{Configurator}, which retrieves candidate meta-strategies from the \textit{Knowledge Base}, selects one to constrain subsequent task-level reasoning, and defines evaluation rules for later reflection in stage 3.

\paragraph{Configurator}
Given a scenario input (domain, goal, background), the Configurator performs two steps: meta-strategy selection and evaluation rule construction.

\textit{Meta-Strategy Selection.} 
The Configurator ranks domain-matched candidates using their historical success counts in the Case Layer of the Knowledge Base (formally defined in stage 3) and selects the best meta-strategy:

\[
M=\argmax_{m\in M(S)} \text{score}(m,S)
\]

\noindent
The selected meta-strategy $M$ is the one with the highest historical success
score in the relevant domain.

\textit{Evaluation Rule Construction.} 
Once $M$ is selected, the Configurator defines evaluation rules $E$ that will be used by the \textit{Evaluator} in stage 3 to judge whether the episode is successful.

Configurator helps MA\textsuperscript{2}P to initiate each persuasion task with explicit planning intent and measurable success criteria.

\subsection{Task-Level Persuading}
After meta-level planning, MA\textsuperscript{2}P executes task-level persuasion via a team of autonomous intelligent agents. 
Guided by the meta-strategy $M$ from the Configurator, stage 2 performs multi-turn reasoning, response generation, and short-term adaptation during interaction.

As shown in Fig.~\ref{fig:solution_overview}, stage 2 includes four agents:
\textit{Perception} extracts belief--desire cues,
\textit{World Model} infers a specific strategy,
\textit{Persuader Agent} realizes it as natural-language utterances, and
\textit{Short-term Memory} maintains dialogue history and intermediate states.

\begin{algorithm}[H]
\caption{MA\textsuperscript{2}P}
\label{alg:ma2p}
\begin{algorithmic}[1]
\Statex \textbf{Input:} scenario $S$ (domain $D(S)$, goal, background); knowledge base $K$
\Statex \textbf{Parameter:} maximum turns $\mathrm{T}_{\max}$
\Statex \textbf{Output:} dialogue $\{(U_t,\tilde{U}_t)\}_{t=1}^{T'}$; updated knowledge base $K'$
\Statex

\vspace{0.2em}
\State \textbf{/* Stage 1*/}
\State $\mathcal{M}\leftarrow \{\, m \in K_{\text{meta}} \mid D(m)=D(S)\,\}$ 
\Comment{retrieve domain-matched meta-strategies}
\ForAll{$m \in \mathcal{M}$}
    \State $\text{score}(m)\leftarrow K_{\text{case}}(m,D(S))$
    \Comment{historical success count in the Case Layer}
\EndFor
\State $M \leftarrow \arg\max_{m\in\mathcal{M}} \text{score}(m)$
\Comment{select the best meta-strategy for this domain}
\State $E \leftarrow \textsc{BuildRules}(M,S)$
\Comment{construct evaluation rules used in Stage 3}

\vspace{0.2em}
\State \textbf{/* Stage 2*/}
\State $H \leftarrow \emptyset$; $\Sigma \leftarrow \emptyset$; $T' \leftarrow \mathrm{T}_{\max}$
\Comment{initialize dialogue history / short-term memory}
\For{$t \leftarrow 1$ \textbf{to} $\mathrm{T}_{\max}$}
    \State $P_t \leftarrow f_{\text{perc}}(H)$
    \Comment{Perception: extract mental-state cues from history}
    \State $\Sigma \leftarrow \{\,H,\;P_t,\;W_{1:t-1}\,\}$
    \Comment{Short-term memory snapshot}
    \State $W_t \leftarrow f_{\text{wm}}(M,\Sigma)$
    \Comment{World model: instantiate $M$ into a concrete strategy}
    \State $U_t \leftarrow f_{\text{pers}}(W_t,H)$
    \Comment{Persuader: realize strategy as an utterance}
    \State $H \leftarrow H \cup \{U_t, \tilde{U}_t\}$
    \Comment{$\tilde{U}_t$ from persuadee}
    \If{$f_{\text{accept}}(H)=1$}
        \State $T' \leftarrow t$; \textbf{break}
    \EndIf
\EndFor

\vspace{0.2em}
\State \textbf{/* Stage 3*/}
\State $R \leftarrow f_{\text{eval}}(E,H)$
\Comment{judge success under rules $E$ using the episode evidence}
\If{$R=1$}
    \State $K_{\text{case}}(M,D(S))+1$
    \Comment{increment domain--strategy success count}
\EndIf
\State $K' \leftarrow \textsc{KBUpdate}(K,M,S,R)$
\Comment{write back the episode summary}
\State \Return $\{(U_t,\tilde{U}_t)\}_{t=1}^{T'},\, K'$
\end{algorithmic}
\end{algorithm}

\subsubsection{Perception}
\textit{Perception} extracts explicit signals and infers latent mental cues (beliefs, desires, concerns) from the dialogue history, which are then provided to the \textit{World Model} for adaptive strategy generation.

\[
P_t = f_{\text{perc}}(H_t)
\]

\noindent Where $H_t$ denotes the dialogue history at turn $t$, and $P_t$ represents the explicit and implicit mental-state cues (beliefs, desires, latent concerns) extracted by the Perception module.

\subsubsection{World Model}
\textit{World Model} is the reasoning core that combines the meta-strategy $M$ with the short-term memory $\Sigma_t$ to infer a specific, actionable persuasion strategy for the next turn.

\[
W_t = f_{\text{wm}}(M,\Sigma_t)
\]

\noindent Where $M$ is the selected meta-strategy from stage 1, 
and $\Sigma_t$ is the short-term memory at turn $t$ containing dialogue history, inferred states, and previous strategies. 
$W_t$ denotes the specific persuasion strategy inferred by the World Model.

It outputs a \textit{specific strategy} aligned with both conversation context and the persuadee's inferred state (e.g., under ``Authority'' with financial concerns, emphasizing insurance coverage or expert endorsement), and passes it to the \textit{Persuader Agent} for realization.

\subsubsection{Persuader Agent}
\textit{Persuader Agent} converts the specific strategy into a coherent utterance conditioned on the dialogue history.

\[
U_t = f_{\text{pers}}(W_t,\; H_t)
\]

\noindent Where $U_t$ denotes the natural-language utterance generated by the Persuader Agent based on the specific strategy $W_t$ and the dialogue history $H_t$.

\[
H_{t+1} = H_t \cup \{\,U_t\,\}
\]
\noindent Where $H_{t+1}$ represents the updated dialogue history after appending the newly generated utterance.

\subsubsection{Short-term Memory}
\textit{Short-term Memory} maintains a shared snapshot of the episode by storing dialogue history, perception outputs, and previously generated strategies.

\[
\Sigma_t = \{\, H_t,\; P_t,\; W_{1:t-1} \,\}
\]
\noindent Where $\Sigma_t$ aggregates three kinds of information at turn $t$: 
the dialogue history $H_t$, inferred mental-state cues $P_t$, and previously generated specific strategies $W_{1:t-1}$.

\[
\Sigma_{t+1} = \text{update}(\Sigma_t, H_{t+1})
\]
\noindent Where the function $\text{update}(\cdot)$ integrates the updated dialogue history $H_{t+1}$ (including both $U_t$ and $\tilde{U}_t$) into memory, producing the next short-term state $\Sigma_{t+1}$.

\subsection{Knowledge Updating}
After each episode, MA\textsuperscript{2}P reflects on the outcome and updates the Knowledge Base to improve future meta-level judging (stage~1). 
Stage~3 comprises (i) a \textit{Knowledge Base} that stores reusable strategy evidence, (ii) an \textit{Evaluator} that checks success under the evaluation rules, and (iii) a \textit{KB-Updater} that writes successful patterns back to the KB.

\subsubsection{Knowledge Base Structure}
As illustrated in Fig.~\ref{fig:solution_overview}, Knowledge Base is organized as a three-layer tree. 
\paragraph{Meta-Strategy Layer} stores high-level persuasion strategies grounded in Cialdini's seven influence principles, e.g., \textit{Authority}, \textit{Commitment and Consistency}, and \textit{Social Proof} \citep{cialdini2021influence};

\paragraph{Domain Layer} groups domains where each strategy has been applied (e.g., \textit{Education}, \textit{Health}, \textit{Finance}); 

\paragraph{Case Layer} records aggregated successful outcomes as counts for each domain--strategy pair.

\[
M(S)=\{\, m \mid m \in K_{\text{meta}} \,\land\, \text{D}(m)=\text{D}(S) \,\}
\]

\[
\text{score}(m,S)=K_{\text{case}}(m,\text{D}(S))
\]

\noindent
Here, $S$ is the given scenario; 
$K_{\text{meta}}$ denotes the meta-strategy layer in the knowledge base $K$; 
$M(S)$ contains all meta-strategies applicable to the domain of $S$.
The function $K_{\text{case}}$ returns the number of successful persuasion
cases associated with meta-strategy $m$ in the domain of scenario $S$,
providing a data-driven measure of prior effectiveness.

\subsubsection{Evaluator}
The \textit{Evaluator} determines whether the episode satisfies the evaluation rules $E$ from stage 1, using the final short-term memory $\Sigma_T$ as evidence.

\[
R = f_{\text{eval}}(E,\Sigma_T)
\]
\noindent Here, $R \in \{0,1\}$ denotes whether the persuasion succeeds, 
$E$ is the evaluation rule set produced in stage 1, 
and $\Sigma_T$ is the final short-term memory at the last dialogue turn $T$.

\subsubsection{Knowledge Base Updating}
If Evaluator confirms success, \textit{Knowledge Base Updating} module increments the Case Layer count for the selected meta-strategy $M$ under the scenario domain $\text{D}(S)$, thereby accumulating evidence of effective domain--strategy pairs over time.

\[
K_{\text{case}}(M, \text{D}(S)) \leftarrow 
K_{\text{case}}(M, \text{D}(S)) + 1
\]
\noindent This operation increments the success-count entry in the Case Layer 
for the selected meta-strategy $M$ under the scenario domain $\text{D}(S)$.

\[
K' = \text{update}(K, M, S, R)
\]
\noindent Where $K'$ is the updated knowledge base after integrating the new experience, 
and $K$ is the previous knowledge base before reflection.

\begin{table*}[t]
\centering
\caption{Main results comparing base LLM persuaders with and without our plug-and-play MA\textsuperscript{2}P. Higher is better for \textbf{Success}, \textbf{Persuasive}, \textbf{Logic}, and \textbf{Helpful}; lower is better for \textbf{Range}, \textbf{SD}, and \textbf{Avg\_Turn}. Green indicates improvements and red indicates degradations relative to the base model.}
\label{tab:main_results}
\small
\setlength{\tabcolsep}{4pt}
\renewcommand{\arraystretch}{1.05}
\resizebox{\textwidth}{!}{%
\begin{tabular}{lccccccc}
\toprule
Model & Success ($\uparrow$) & Persuasive ($\uparrow$) & Logic ($\uparrow$) & Helpful ($\uparrow$) & Range ($\downarrow$) & SD ($\downarrow$) & Avg\_Turn ($\downarrow$) \\
\midrule
gpt-4o-mini & 0.45 & 6.46 & 7.82 & 7.58 & 0.450 & 0.104 & 2.94 \\
+ MA\textsuperscript{2}P       & 0.79 \textcolor{ForestGreen}{(+0.34)} & 7.16 \textcolor{ForestGreen}{(+0.70)} & 8.08 \textcolor{ForestGreen}{(+0.26)} & 8.07 \textcolor{ForestGreen}{(+0.49)} & 0.400 \textcolor{ForestGreen}{(-0.05)} & 0.107 \textcolor{BrickRed}{(+)} & 1.86 \textcolor{ForestGreen}{(-1.08)} \\
\midrule
gpt-4o      & 0.46 & 6.45 & 7.87 & 7.70 & 0.500 & 0.114 & 3.03 \\
+ MA\textsuperscript{2}P       & 0.75 \textcolor{ForestGreen}{(+0.29)} & 6.99 \textcolor{ForestGreen}{(+0.54)} & 7.96 \textcolor{ForestGreen}{(+0.09)} & 7.94 \textcolor{ForestGreen}{(+0.24)} & 0.488 \textcolor{ForestGreen}{(-0.01)} & 0.109 \textcolor{ForestGreen}{(-0.01)} & 2.00 \textcolor{ForestGreen}{(-1.03)} \\
\midrule
gpt-5-mini  & 0.51 & 6.40 & 7.81 & 7.55 & 0.716 & 0.145 & 2.66 \\
+ MA\textsuperscript{2}P       & 0.72 \textcolor{ForestGreen}{(+0.21)} & 7.15 \textcolor{ForestGreen}{(+0.75)} & 8.28 \textcolor{ForestGreen}{(+0.47)} & 8.27 \textcolor{ForestGreen}{(+0.72)} & 0.491 \textcolor{ForestGreen}{(-0.23)} & 0.115 \textcolor{ForestGreen}{(-0.03)} & 1.60 \textcolor{ForestGreen}{(-1.06)} \\
\midrule
gemini-2.5-flash & 0.46 & 6.48 & 7.85 & 7.62 & 0.400 & 0.113 & 3.27 \\
+ MA\textsuperscript{2}P            & 0.66 \textcolor{ForestGreen}{(+0.20)} & 6.67 \textcolor{ForestGreen}{(+0.19)} & 7.78 \textcolor{BrickRed}{(-0.07)} & 7.44 \textcolor{BrickRed}{(-0.18)} & 0.381 \textcolor{ForestGreen}{(-0.02)} & 0.092 \textcolor{ForestGreen}{(-0.02)} & 2.08 \textcolor{ForestGreen}{(-1.19)} \\
\midrule
deepseek-v3  & 0.53 & 6.98 & 8.06 & 7.84 & 0.412 & 0.107 & 3.05 \\
+ MA\textsuperscript{2}P       & 0.80 \textcolor{ForestGreen}{(+0.27)} & 7.58 \textcolor{ForestGreen}{(+0.60)} & 8.51 \textcolor{ForestGreen}{(+0.45)} & 8.42 \textcolor{ForestGreen}{(+0.58)} & 0.400 \textcolor{ForestGreen}{(-0.01)} & 0.088 \textcolor{ForestGreen}{(-0.02)} & 1.82 \textcolor{ForestGreen}{(-1.23)} \\
\bottomrule
\end{tabular}%
}
\end{table*}

\section{Experiment}

In this section, we evaluate MA\textsuperscript{2}P by comparing multiple base LLMs against their MA\textsuperscript{2}P-augmented counterparts under a plug-and-play, training-free setting. We further conduct ablation studies to isolate the contribution of key components, and report both LLM-based and human A/B preference evaluations. Across these complementary protocols, the results consistently demonstrate the effectiveness of MA\textsuperscript{2}P.

\subsection{Main Results}
In this subsection, we compare each base LLM persuader with its MA\textsuperscript{2}P-augmented counterpart.

\paragraph{Setup.}
We evaluate five base models: \texttt{gpt-4o-mini}, \texttt{gpt-4o}, \texttt{gpt-5-mini}, \texttt{gemini-2.5-flash}, and \texttt{deepseek-v3}.
For each model, we compare a baseline next-turn persuader with its MA\textsuperscript{2}P-augmented counterpart. MA\textsuperscript{2}P is plug-and-play and training-free, so the same framework is applied on top of each base model.
Experiments are conducted on the CToMPersu dataset \citep{zhang2025persuasiondoubleblindmultidomaindialogue}, using the official test split of 525 instances, where each scenario includes annotated persuadee mental states (belief and desire) about the original intention (when available) and the target action.
To control for persuadee variability, we use a fixed \texttt{gpt-4o-mini} agent as the persuadee in all settings; it is conditioned on the scenario and mental-state annotations.

\paragraph{Metrics.}
Let $\mathcal{D}$ be the test set, $|\mathcal{D}|=N$. For each instance $i$, we generate a dialogue of at most $T_{\max}=4$ turns. Let $J(\cdot)\in\{0,1\}$ be the success judge on a dialogue prefix, and let $Persu(\cdot),L(\cdot),H(\cdot)\in\{1,\ldots,10\}$ be the LLM scores for \textit{Persuasive}, \textit{Logic}, and \textit{Helpful}. All judges use \texttt{gpt-4o-mini}.

\textbf{Success.} A test case is successful if the judge returns \texttt{True} at any turn $t\le T_{\max}$.
\[
\mathrm{Success}=\frac{1}{N}\sum_{i=1}^{N}\mathbf{1}\!\left[\max_{t\le T_{\max}} J\!\left(d_i^{\le t}\right)=1\right].
\]

\textbf{Persuasive.} Average persuasive-strength score on a 1--10 scale.
\[
\mathrm{Persuasive}=\frac{1}{N}\sum_{i=1}^{N} Persu(d_i).
\]

\textbf{Logic.} Average logical-coherence score on a 1--10 scale.
\[
\mathrm{Logic}=\frac{1}{N}\sum_{i=1}^{N} L(d_i).
\]

\textbf{Helpful.} Average helpfulness score on a 1--10 scale.
\[
\mathrm{Helpful}=\frac{1}{N}\sum_{i=1}^{N} H(d_i).
\]

\textbf{Range.}
Let $\mathcal{G}$ denote the 35 domains. For each $g\in\mathcal{G}$, let $\mathrm{Succ}_g$ be the success rate computed on test cases from domain $g$.
\[
\mathrm{Range}=\max_{g\in\mathcal{G}}\mathrm{Succ}_g-\min_{g\in\mathcal{G}}\mathrm{Succ}_g.
\]

\textbf{SD.} Standard deviation of domain-wise success rates.
\[
\mathrm{SD}=\sqrt{\frac{1}{|\mathcal{G}|}\sum_{g\in\mathcal{G}}\left(\mathrm{Succ}_g-\overline{\mathrm{Succ}}\right)^2}
\]

\textbf{Avg\_Turn.}
Let $\tau_i$ be the first turn $t\le T_{\max}$ where $J(d_i^{\le t})=1$, or $\tau_i=T_{\max}$ if never. Then
\[
\mathrm{Avg\_Turn}=\frac{1}{N}\sum_{i=1}^{N}\tau_i.
\]

\paragraph{Results.}
Table~\ref{tab:main_results} shows that MA\textsuperscript{2}P improves over the corresponding baselines on almost all metrics across the five base models. The main exceptions occur for \texttt{gemini-2.5-flash}, where Logic and Helpful slightly decrease; however, MA\textsuperscript{2}P still delivers a large gain in \textbf{Success} (over $+0.20$ absolute), suggesting that the framework remains effective for achieving the persuasion goal even when a few quality metrics do not improve. We also observe a minor increase in SD on \texttt{gpt-4o-mini}, while Range still decreases. Overall, MA\textsuperscript{2}P consistently boosts persuasion success and dialogue quality, and generally reduces cross-domain dispersion (Range/SD), indicating stronger robustness and generalization.

\begin{table}[t]
\centering
\caption{Ablation results comparing the base LLM, an autonomous-agent system without meta-cognitive augmentation (\texttt{+Auto}), and the full MA\textsuperscript{2}P system (\texttt{+MA\textsuperscript{2}P}). Green/red indicate improvements/degradations over the base model.}

\label{tab:ablation}
\footnotesize
\renewcommand{\arraystretch}{1.05}

\setlength{\tabcolsep}{0pt}

\begin{tabular}{l@{\hspace{10pt}}c@{\hspace{10pt}}c@{\hspace{10pt}}c}
\toprule
Model & Success ($\uparrow$) & Range ($\downarrow$) & SD ($\downarrow$) \\
\midrule
4o-mini & 0.45 & 0.450 & 0.104 \\
\hspace{0.6em}+ Auto
& 0.66 \textcolor{ForestGreen}{(+0.21)}
& 0.530 \textcolor{BrickRed}{(+0.08)}
& 0.118 \textcolor{BrickRed}{(+)} \\
\hspace{0.6em}+ MA\textsuperscript{2}P
& 0.79 \textcolor{ForestGreen}{(+0.34)}
& 0.400 \textcolor{ForestGreen}{(-0.05)}
& 0.107 \textcolor{BrickRed}{(+)} \\
\midrule
4o & 0.46 & 0.500 & 0.114 \\
\hspace{0.6em}+ Auto
& 0.68 \textcolor{ForestGreen}{(+0.22)}
& 0.458 \textcolor{ForestGreen}{(-0.04)}
& 0.120 \textcolor{BrickRed}{(+0.01)} \\
\hspace{0.6em}+ MA\textsuperscript{2}P
& 0.75 \textcolor{ForestGreen}{(+0.29)}
& 0.488 \textcolor{ForestGreen}{(-0.01)}
& 0.109 \textcolor{ForestGreen}{(-0.01)} \\
\bottomrule
\end{tabular}
\end{table}

\subsection{Ablation Study}
We additionally report a variant that uses a system of autonomous intelligent agents without meta-cognitive augmentation (\texttt{+Auto}).
Since our meta-cognition is implemented as an explicit enhancement to the Configurator inside this agent system, it cannot be meaningfully evaluated as a standalone component.

As shown in Table~\ref{tab:ablation}, (\texttt{+Auto}) already yields a clear improvement in persuasion success over the base model.
However, it exhibits weaker cross-domain robustness, reflected by larger domain dispersion (e.g., higher \textbf{Range} and/or \textbf{SD}).
This suggests that (\texttt{+Auto}) tends to amplify performance in domains where the base model is already strong, rather than uniformly improving all domains.
By contrast, the full MA\textsuperscript{2}P system (\texttt{+MA\textsuperscript{2}P}) further increases \textbf{Success} while simultaneously reducing cross-domain variance, indicating that meta-cognitive augmentation helps the system generalize more consistently across domains.

\paragraph{Warm-up episodes and knowledge base size.}
Since MA\textsuperscript{2}P relies on the knowledge base for meta-strategy selection, we further study how the number of warm-up episodes affects performance. Specifically, we vary the initial knowledge base size $K$ and evaluate the resulting system on \texttt{gpt-4o-mini}, while keeping the rest of the setup unchanged. Table~\ref{tab:kb_size} reports the results.

\begin{table}[h]
\centering
\small
\caption{Effect of the initial knowledge base size $K$ on MA\textsuperscript{2}P (\texttt{gpt-4o-mini}).}
\begin{tabular}{lccc}
\toprule
\textbf{Model (4o-mini)} & \textbf{Success ($\uparrow$)} & \textbf{Range ($\downarrow$)} & \textbf{SD ($\downarrow$)} \\
\midrule
$K=0$   & 0.66 & 0.53 & 0.118 \\
$K=100$ & 0.73 & 0.44 & 0.107 \\
$K=300$ & 0.76 & 0.51 & 0.111 \\
$K=500$ & 0.79 & 0.40 & 0.107 \\
\bottomrule
\end{tabular}
\label{tab:kb_size}
\end{table}

The results show that MA\textsuperscript{2}P remains effective even under cold-start conditions ($K=0$), but a small number of warm-up episodes already leads to clear gains. In particular, with $K=100$, the system achieves substantial improvements in both persuasion success and cross-domain robustness. Increasing the knowledge base further generally brings additional benefits, and the best overall performance is obtained at $K=500$, which is the setting adopted in our main experiments. These results suggest that MA\textsuperscript{2}P does not require a large amount of warm-up data to be useful, while a larger knowledge base can further improve stability and effectiveness.

\subsection{LLM and Human Preference Study}
We conduct an A/B preference study with an LLM judge and human evaluators to compare baseline persuaders against their MA\textsuperscript{2}P-augmented counterparts.
For each test instance, evaluators are presented with two anonymized dialogue outputs (the order is randomized and the system identity is hidden), and they provide a ternary judgment over the \emph{second} output relative to the \emph{first}: \textbf{better} (win), \textbf{comparable} (tie), or \textbf{worse} (lose).

\paragraph{LLM judge.}
We use \texttt{gpt-4o-mini} as the LLM evaluator.
It reads the two anonymized dialogues in full and returns a single win/tie/lose decision under the same blind setting (i.e., without knowing which system produced which output).

\paragraph{Human evaluation.}
We additionally recruit two M.Sc.\ students in computer science as human evaluators.
To reduce annotation cost, we randomly sample 400 test instances for the human study.

\begin{figure}[t]
  \centering
  \includegraphics[width=\linewidth]{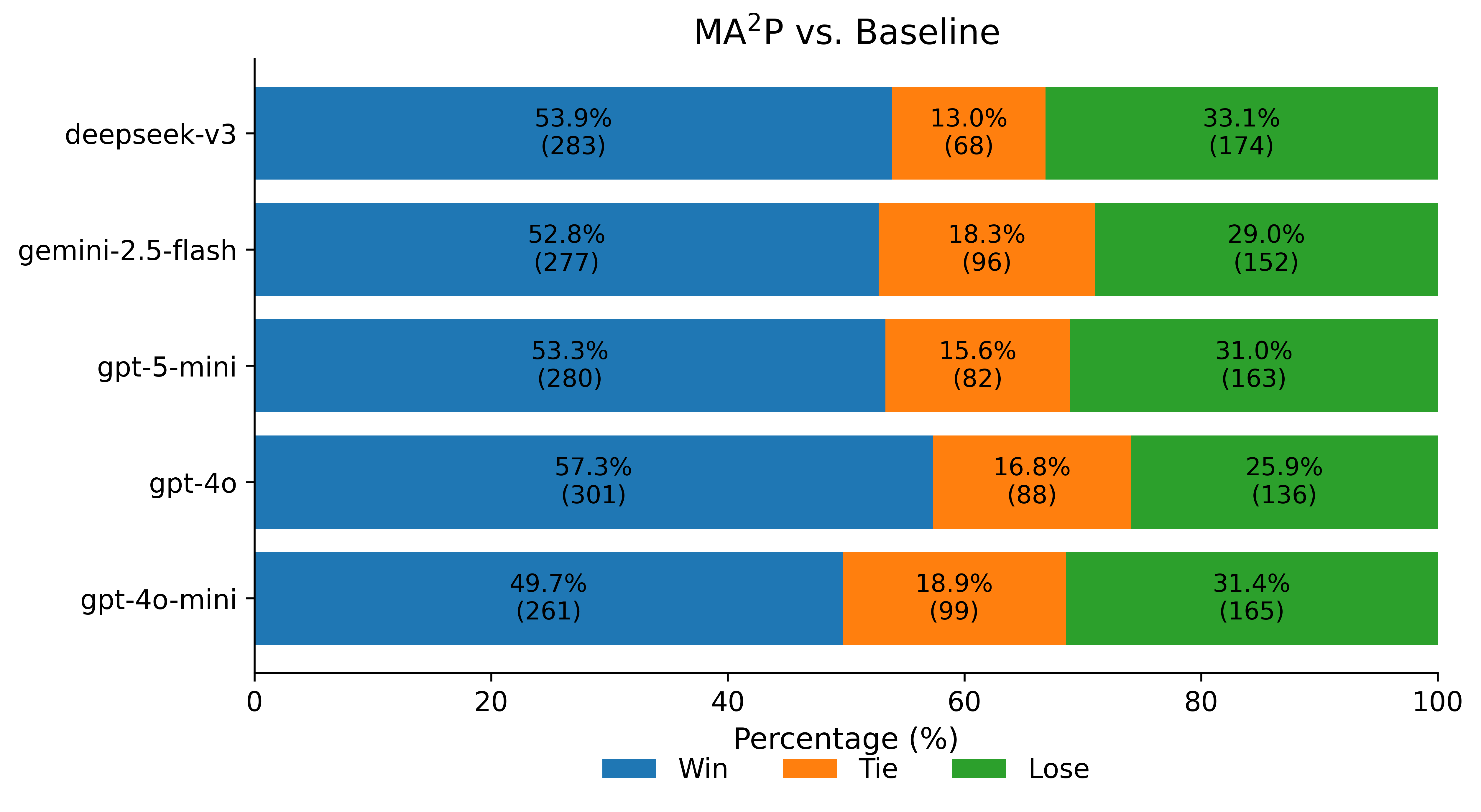}\vspace{0.1cm}
  \includegraphics[width=\linewidth]{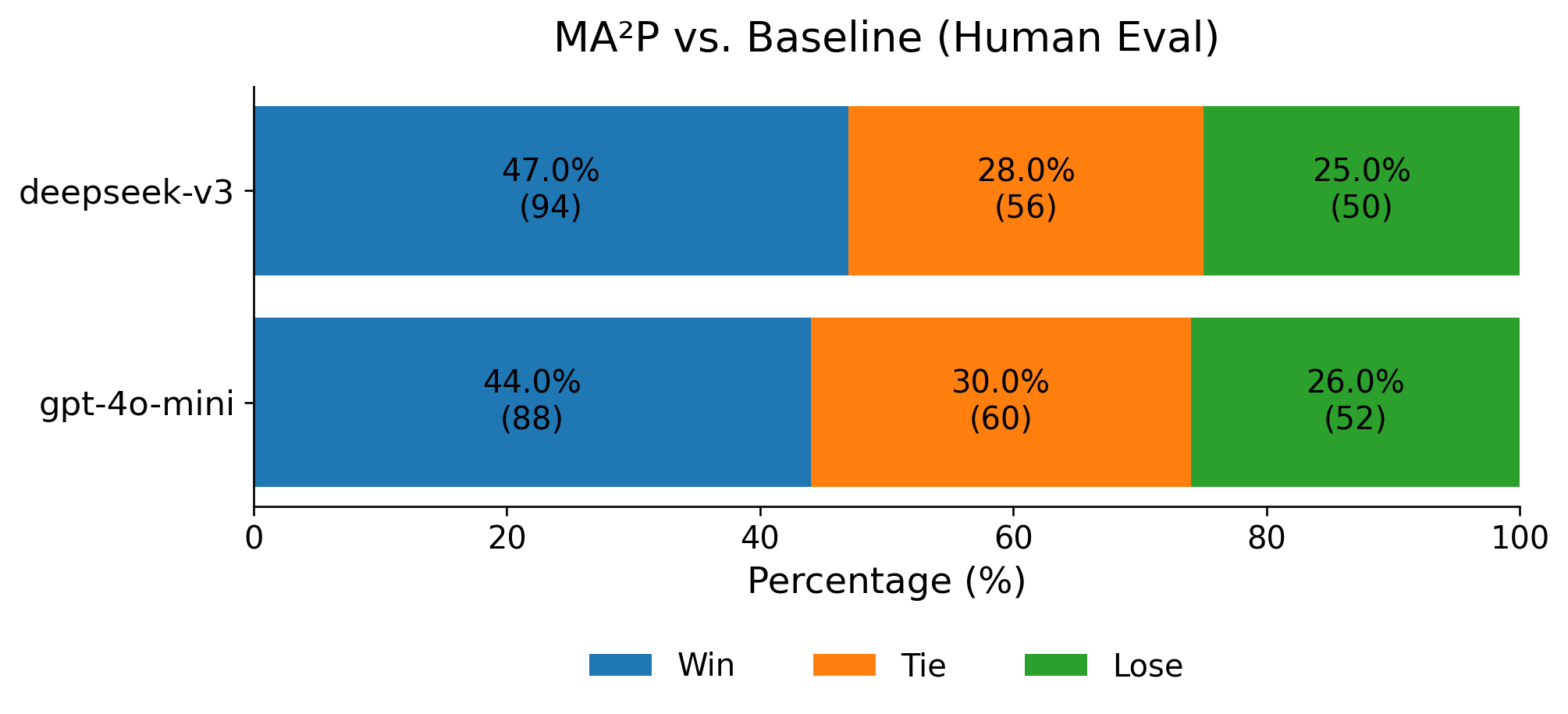}
  \caption{A/B preference results (win/tie/lose) comparing baseline persuaders with their MA\textsuperscript{2}P-augmented counterparts under LLM-based and human evaluation.}
  \label{fig:llm_compare}
  \vspace{-0.2cm}
\end{figure}

As shown in Figure~\ref{fig:llm_compare}, MA\textsuperscript{2}P is consistently preferred over the baseline in both evaluation settings, indicating that MA\textsuperscript{2}P produces more convincing and higher-quality persuasive responses in practice.

\paragraph{LLM--human agreement.}
To quantify the agreement between the LLM judge and human evaluators on the ternary preference labels, we compute the \emph{weighted Cohen's kappa} $\kappa_w$ \citep{cohen1968weighted}.
We treat the three labels as ordinal categories (lose $<$ tie $<$ win) and use quadratic weighting.
Since our human evaluation is conducted on two disjoint subsets (200 instances for \texttt{deepseek-v3} and 200 instances for \texttt{gpt-4o-mini}), we compute $\kappa_w$ separately for each subset (LLM vs.\ the corresponding human evaluator) and report their average.
In our study, we obtain $\kappa_w=\text{0.549}$, indicating moderate agreement between the LLM judge and human evaluators.
We also observe that human evaluators assign \texttt{tie} more frequently when the two responses are close, reflecting the inherent subjectivity of preference judgments.
Nevertheless, the overall trend is consistent: MA\textsuperscript{2}P is preferred over the baseline under both LLM and human evaluation.

To complement the preference results above, we further provide in Appendix~\ref{sec:interactive_qualitative} a small-scale interactive qualitative example from blind human--system conversations, which offers illustrative evidence beyond the simulation setting.

\subsection{Case Study}
\begin{figure}[t]
    \centering
    \includegraphics[width=\linewidth]{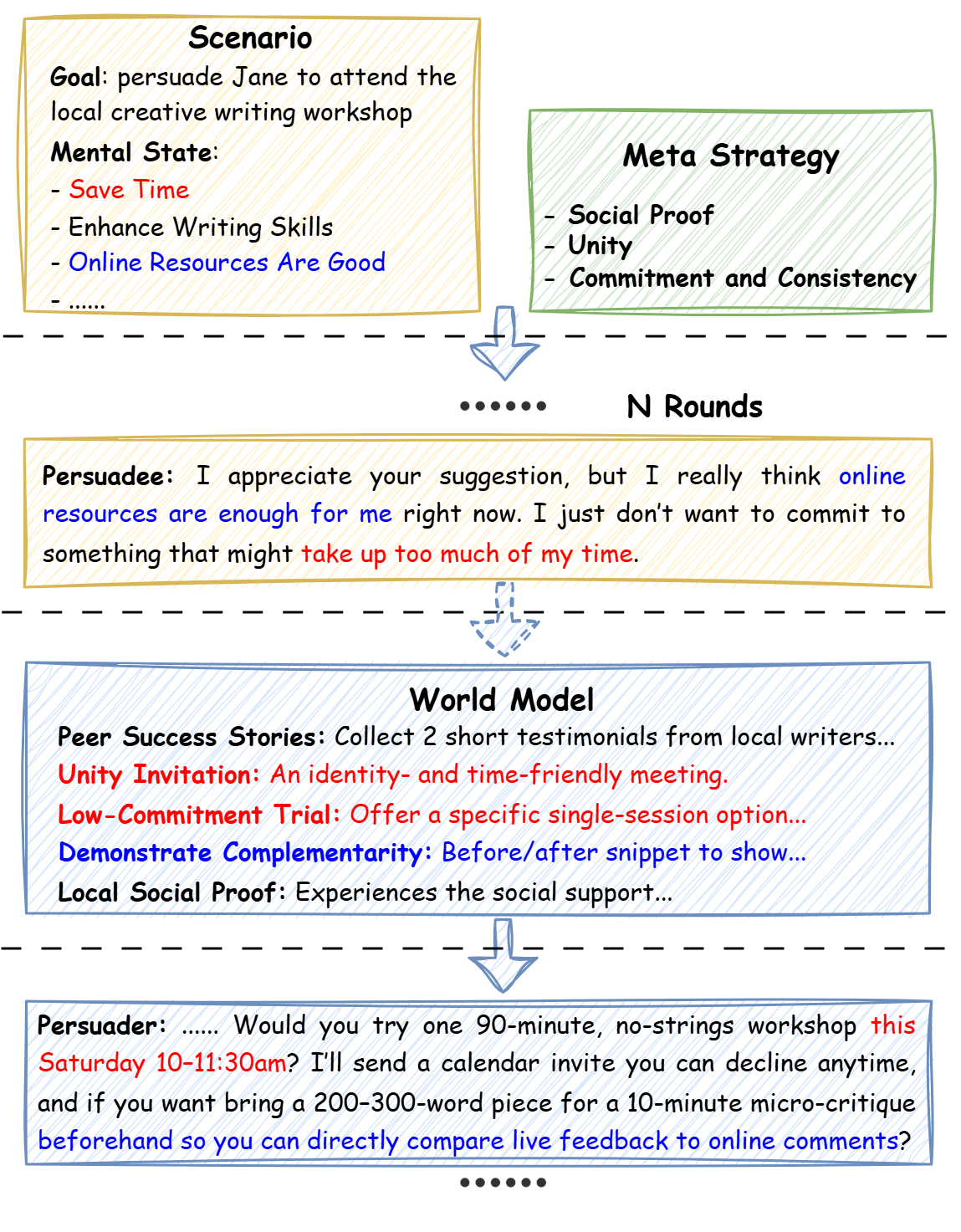}
    \caption{A representative MA\textsuperscript{2}P interaction round.}
    \label{fig:case-study}
    \vspace{-0.3cm}
\end{figure}

Figure~\ref{fig:case-study} shows a representative interaction round. 
MA\textsuperscript{2}P first identifies the persuadee's key mental-state cues (e.g., wanting to \emph{save time} and believing \emph{online resources are sufficient}).

Conditioned on the selected meta strategies, the world model then translates them into concrete, executable persuasion strategies (e.g., a \emph{low-commitment trial} and \emph{demonstrating complementarity} via an online-vs.-in-person comparison). 
Finally, the persuader produces a targeted utterance by proposing a short weekend session to address the time constraint and asking the persuadee to bring a brief writing sample for a live-vs.-online feedback comparison, directly grounding the response in the inferred concerns.

\section{Related Work}

\paragraph{Persuasion with large language models.}
Recent work on LLM persuasion includes: (i) \textbf{surveys and empirical analyses} of persuasive behaviors and risks \citep{rogiers2024persuasionlargelanguagemodels,SINGH2025100197}, spanning public health, consumer decisions, and political contexts \citep{altay2023information,chen2023would,potter-etal-2024-hidden}; (ii) \textbf{user-aware persuasion} that explicitly models user states and adapts psychological strategies beyond surface fluency \citep{dewynter2025linecomprehensionpersuasionllms,ju2025adaptivepsychologicalpersuasionlarge}. Meanwhile, datasets have progressed from single-domain donation dialogues \citep{wang-etal-2019-persuasion} to multi-domain benchmarks \citep{jin-etal-2024-persuading}, and further to double-blind settings exposing belief and desire, increasing the need for state-conditioned persuasion \citep{zhang2025persuasiondoubleblindmultidomaindialogue}. (iii) \textbf{Strategic formulations} frame persuasion as sequential decision making (e.g., persuasion games and off-policy evaluation), emphasizing planful and strategy-aware action selection \citep{karande-etal-2024-persuasion,10.1162/TACL.a.16,cheng2025strategicpersuasionlanguagemodels,jin-etal-2023-joint}. (iv) \textbf{Agentic directions} build modular or multi-agent persuaders for multi-turn interaction, including simulation-based data generation and zero-shot strategy-plus-retrieval frameworks \citep{li-etal-2025-mads,furumai-etal-2024-zero}; related work also studies credibility-aware persuasion and multi-agent reasoning dynamics \citep{qin-etal-2024-beyond,saenger-etal-2024-autopersuade,zhao2025disagreementsreasoningmodelsthinking}.

\paragraph{Meta-cognition for large language models.}
We view metacognition as ``knowing about knowing'' and model it as a controller that decides when to plan, verify, revise, or stop \citep{Metcalfe1994MetacognitionK}. Prior work instantiates this meta-level via: (i) \textbf{prompt/interface} methods for self-questioning, rule interpretation, and transfer \citep{wang-zhao-2024-metacognitive,khan2025literalliberalmetapromptingframework,gong2025tuning,gu2025metawriter,xiao2025promptmii}; (ii) \textbf{reasoning-process} methods that model meta-reasoning signals to improve robustness, align intermediate traces with factuality and final answers, and enable controllable thinking \citep{wang-etal-2024-meta,wang2025mralignmetareasoninginformedfactuality,ha2025ahamomentscontrollablethinking}; (iii) \textbf{agent orchestration} that uses meta-control for tool use and planning, including meta-plans, multi-agent meta-thinking, reflection retrieval, and self-improving judges \citep{li-etal-2025-adaptive,xiong-etal-2025-mpo,wan2025rema,wang-etal-2025-meta,wu-etal-2025-meta}; and (iv) \textbf{evaluation/applications} that argue metacognition should be measured separately from memorization or raw accuracy, and apply it to role-play control, multimodal knowledge editing, rumor detection, and evidence-sensitive medical RAG \citep{li-etal-2024-meta,li2024metacognitiveanalysisevaluatingdeclarative,zhang-etal-2025-remembering,qin-etal-2025-r,fan2025metacognitiveknowledgeeditingmultimodal,he-etal-2025-gcml,sun2025metaragmetaanalysisinspiredevidencererankingmethod}.

\section{Conclusion}
We proposed MA\textsuperscript{2}P, a plug-and-play, training-free meta-cognitive agent framework for complex persuasive dialogue that converts inferred mental-state cues into strategy-grounded actions and reduces cross-domain variance via meta-strategy selection from a structured knowledge base. Experiments across five base LLM persuaders and both LLM and human evaluations show consistent improvements in persuasion success and overall response quality. The appendix provides our prompt designs and additional experimental analyses.

\section*{Limitations}

Due to the inherent difficulty of evaluating open-ended generation, most of our automatic metrics rely on an LLM evaluator; to reduce subjectivity, we use explicit and consistent rubrics for each dimension. Our human preference study is constrained by time and budget, resulting in a small number of annotators and a limited sample size, but it still provides supportive evidence by exhibiting the same preference trend as the LLM-based evaluation. In addition, applying MA\textsuperscript{2}P to a new domain requires a warm-up phase: we need to run a small number of initial episodes to seed the knowledge base with domain-relevant cases for meta-strategy selection. Finally, current persuadee modeling is relatively simple: while we condition the simulated persuadee on belief and desire, we do not model richer traits such as personality. A promising future direction is to develop a standardized \emph{persuadee model} benchmark that does not require dialogue datasets, but evaluates persuasion systems against simulated agents with controllable personalities and belief-update capabilities.

\section*{Ethics Statement}

Our work studies persuasive dialogue generation, which can be beneficial for applications such as education and counseling but may also be misused for manipulation. To mitigate this risk, we focus on research settings with clearly specified user goals and do not target sensitive attributes or political persuasion. All experiments are conducted in an offline simulation environment with an LLM-based persuadee and evaluator; no real users are deceived or influenced, and no personally identifiable information is collected or processed. For the human preference study, participation was voluntary and based on informed consent; annotators evaluated anonymized system outputs, and no sensitive personal data were recorded. We will release our code, prompts, and knowledge base with documentation that highlights appropriate use cases and limitations, and we encourage future work to incorporate additional safeguards and misuse risk assessments when deploying persuasive systems in real-world contexts. We used AI assistants (e.g., ChatGPT) to help with language polishing and drafting/brainstorming during writing. All outputs were reviewed and edited by the authors, who take full responsibility for the content and correctness.

\section*{Acknowledgments}
We thank the anonymous reviewers and area chairs for their valuable comments and suggestions, which helped improve this paper. Also, we thank Shencheng Yao from the International Department of Chengdu No. 7 High School for assistance with early-stage data inspection, preliminary error case analysis, and checking a portion of the initial annotations. We gratefully acknowledge his support during the exploratory stage of this project. This work was supported by the National Natural Science Foundation of China (62176053).

\bibliography{custom}

@article{10.1162/TACL.a.16,
    author = {Shapira, Eilam and Madmon, Omer and Apel, Reut and Tennenholtz, Moshe and Reichart, Roi},
    title = {Human Choice Prediction in Language-based Persuasion Games: Simulation-based Off-Policy Evaluation},
    journal = {Transactions of the Association for Computational Linguistics},
    volume = {13},
    pages = {980-1006},
    year = {2025},
    month = {08},
    issn = {2307-387X},
    doi = {10.1162/TACL.a.16},
    url = {https://doi.org/10.1162/TACL.a.16},
    eprint = {https://direct.mit.edu/tacl/article-pdf/doi/10.1162/TACL.a.16/2549171/tacl.a.16.pdf},
}

@misc{rogiers2024persuasionlargelanguagemodels,
      title={Persuasion with Large Language Models: a Survey}, 
      author={Alexander Rogiers and Sander Noels and Maarten Buyl and Tijl De Bie},
      year={2024},
      eprint={2411.06837},
      archivePrefix={arXiv},
      primaryClass={cs.CL},
      url={https://arxiv.org/abs/2411.06837}, 
}

@misc{cheng2025strategicpersuasionlanguagemodels,
      title={Towards Strategic Persuasion with Language Models}, 
      author={Zirui Cheng and Jiaxuan You},
      year={2025},
      eprint={2509.22989},
      archivePrefix={arXiv},
      primaryClass={cs.AI},
      url={https://arxiv.org/abs/2509.22989}, 
}

@article{SINGH2025100197,
title = {The influence of persuasive techniques on large language models: A scenario-based study},
journal = {Computers in Human Behavior: Artificial Humans},
volume = {6},
pages = {100197},
year = {2025},
issn = {2949-8821},
doi = {https://doi.org/10.1016/j.chbah.2025.100197},
url = {https://www.sciencedirect.com/science/article/pii/S2949882125000817},
author = {Sonali Uttam Singh and Akbar Siami Namin}
}

@misc{zhao2025disagreementsreasoningmodelsthinking,
      title={Disagreements in Reasoning: How a Model's Thinking Process Dictates Persuasion in Multi-Agent Systems}, 
      author={Haodong Zhao and Jidong Li and Zhaomin Wu and Tianjie Ju and Zhuosheng Zhang and Bingsheng He and Gongshen Liu},
      year={2025},
      eprint={2509.21054},
      archivePrefix={arXiv},
      primaryClass={cs.AI},
      url={https://arxiv.org/abs/2509.21054}, 
}

@inproceedings{li-etal-2025-mads,
    title = "{MADS}: Multi-Agent Dialogue Simulation for Diverse Persuasion Data Generation",
    author = "Li, Mingjin  and
      Liu, Yu  and
      Liu, Huayi  and
      Ye, Xiang  and
      Jiang, Chao  and
      Zhang, Hongguang  and
      Ruan, Yu",
    editor = "Potdar, Saloni  and
      Rojas-Barahona, Lina  and
      Montella, Sebastien",
    booktitle = "Proceedings of the 2025 Conference on Empirical Methods in Natural Language Processing: Industry Track",
    month = nov,
    year = "2025",
    address = "Suzhou (China)",
    publisher = "Association for Computational Linguistics",
    url = "https://aclanthology.org/2025.emnlp-industry.26/",
    doi = "10.18653/v1/2025.emnlp-industry.26",
    pages = "399--415",
    ISBN = "979-8-89176-333-3"
}

@inproceedings{karande-etal-2024-persuasion,
    title = "Persuasion Games with Large Language Models",
    author = "Karande, Shirish  and
      V, Santhosh  and
      Bhatia, Yash",
    editor = "Lalitha Devi, Sobha  and
      Arora, Karunesh",
    booktitle = "Proceedings of the 21st International Conference on Natural Language Processing (ICON)",
    month = dec,
    year = "2024",
    address = "AU-KBC Research Centre, Chennai, India",
    publisher = "NLP Association of India (NLPAI)",
    url = "https://aclanthology.org/2024.icon-1.67/",
    pages = "576--582"
}

@misc{zhang2025persuasiondoubleblindmultidomaindialogue,
      title={Persuasion Should be Double-Blind: A Multi-Domain Dialogue Dataset With Faithfulness Based on Causal Theory of Mind}, 
      author={Dingyi Zhang and Deyu Zhou},
      year={2025},
      eprint={2502.21297},
      archivePrefix={arXiv},
      primaryClass={cs.CL},
      url={https://arxiv.org/abs/2502.21297}, 
}

@inproceedings{furumai-etal-2024-zero,
    title = "Zero-shot Persuasive Chatbots with {LLM}-Generated Strategies and Information Retrieval",
    author = "Furumai, Kazuaki  and
      Legaspi, Roberto  and
      Romero, Julio Cesar Vizcarra  and
      Yamazaki, Yudai  and
      Nishimura, Yasutaka  and
      Semnani, Sina  and
      Ikeda, Kazushi  and
      Shi, Weiyan  and
      Lam, Monica",
    editor = "Al-Onaizan, Yaser  and
      Bansal, Mohit  and
      Chen, Yun-Nung",
    booktitle = "Findings of the Association for Computational Linguistics: EMNLP 2024",
    month = nov,
    year = "2024",
    address = "Miami, Florida, USA",
    publisher = "Association for Computational Linguistics",
    url = "https://aclanthology.org/2024.findings-emnlp.656/",
    doi = "10.18653/v1/2024.findings-emnlp.656",
    pages = "11224--11249"
}

@inproceedings{qin-etal-2024-beyond,
    title = "Beyond Persuasion: Towards Conversational Recommender System with Credible Explanations",
    author = "Qin, Peixin  and
      Huang, Chen  and
      Deng, Yang  and
      Lei, Wenqiang  and
      Chua, Tat-Seng",
    editor = "Al-Onaizan, Yaser  and
      Bansal, Mohit  and
      Chen, Yun-Nung",
    booktitle = "Findings of the Association for Computational Linguistics: EMNLP 2024",
    month = nov,
    year = "2024",
    address = "Miami, Florida, USA",
    publisher = "Association for Computational Linguistics",
    url = "https://aclanthology.org/2024.findings-emnlp.247/",
    doi = "10.18653/v1/2024.findings-emnlp.247",
    pages = "4264--4282"
}

@misc{ju2025adaptivepsychologicalpersuasionlarge,
      title={On the Adaptive Psychological Persuasion of Large Language Models}, 
      author={Tianjie Ju and Yujia Chen and Hao Fei and Mong-Li Lee and Wynne Hsu and Pengzhou Cheng and Zongru Wu and Zhuosheng Zhang and Gongshen Liu},
      year={2025},
      eprint={2506.06800},
      archivePrefix={arXiv},
      primaryClass={cs.CL},
      url={https://arxiv.org/abs/2506.06800}, 
}

@misc{dewynter2025linecomprehensionpersuasionllms,
      title={The Thin Line Between Comprehension and Persuasion in LLMs}, 
      author={Adrian de Wynter and Tangming Yuan},
      year={2025},
      eprint={2507.01936},
      archivePrefix={arXiv},
      primaryClass={cs.CL},
      url={https://arxiv.org/abs/2507.01936}, 
}

@article{altay2023information,
  title={Information delivered by a chatbot has a positive impact on COVID-19 vaccines attitudes and intentions.},
  author={Altay, Sacha and Hacquin, Anne-Sophie and Chevallier, Coralie and Mercier, Hugo},
  journal={Journal of Experimental Psychology: Applied},
  volume={29},
  number={1},
  pages={52},
  year={2023},
  publisher={American Psychological Association}
}

@inproceedings{potter-etal-2024-hidden,
    title = "Hidden Persuaders: {LLM}s' Political Leaning and Their Influence on Voters",
    author = "Potter, Yujin  and
      Lai, Shiyang  and
      Kim, Junsol  and
      Evans, James  and
      Song, Dawn",
    editor = "Al-Onaizan, Yaser  and
      Bansal, Mohit  and
      Chen, Yun-Nung",
    booktitle = "Proceedings of the 2024 Conference on Empirical Methods in Natural Language Processing",
    month = nov,
    year = "2024",
    address = "Miami, Florida, USA",
    publisher = "Association for Computational Linguistics",
    url = "https://aclanthology.org/2024.emnlp-main.244/",
    doi = "10.18653/v1/2024.emnlp-main.244",
    pages = "4244--4275"
}

@article{chen2023would,
  title={Would an AI chatbot persuade you: an empirical answer from the elaboration likelihood model},
  author={Chen, Qian and Yin, Changqin and Gong, Yeming},
  journal={Information Technology \& People},
  year={2023},
  publisher={Emerald Publishing Limited}
}

@inproceedings{jin-etal-2023-joint,
    title = "Joint Semantic and Strategy Matching for Persuasive Dialogue",
    author = "Jin, Chuhao  and
      Zhu, Yutao  and
      Kong, Lingzhen  and
      Li, Shijie  and
      Zhang, Xiao  and
      Song, Ruihua  and
      Chen, Xu  and
      Chen, Huan  and
      Sun, Yuchong  and
      Chen, Yu  and
      Xu, Jun",
    editor = "Bouamor, Houda  and
      Pino, Juan  and
      Bali, Kalika",
    booktitle = "Findings of the Association for Computational Linguistics: EMNLP 2023",
    month = dec,
    year = "2023",
    address = "Singapore",
    publisher = "Association for Computational Linguistics",
    url = "https://aclanthology.org/2023.findings-emnlp.276/",
    doi = "10.18653/v1/2023.findings-emnlp.276",
    pages = "4187--4197"
}

@inproceedings{wang-etal-2019-persuasion,
    title = "Persuasion for Good: Towards a Personalized Persuasive Dialogue System for Social Good",
    author = "Wang, Xuewei  and
      Shi, Weiyan  and
      Kim, Richard  and
      Oh, Yoojung  and
      Yang, Sijia  and
      Zhang, Jingwen  and
      Yu, Zhou",
    editor = "Korhonen, Anna  and
      Traum, David  and
      M{\`a}rquez, Llu{\'i}s",
    booktitle = "Proceedings of the 57th Annual Meeting of the Association for Computational Linguistics",
    month = jul,
    year = "2019",
    address = "Florence, Italy",
    publisher = "Association for Computational Linguistics",
    url = "https://aclanthology.org/P19-1566/",
    doi = "10.18653/v1/P19-1566",
    pages = "5635--5649"
}

@inproceedings{saenger-etal-2024-autopersuade,
    title = "{A}uto{P}ersuade: A Framework for Evaluating and Explaining Persuasive Arguments",
    author = "Saenger, Till Raphael  and
      Hinck, Musashi  and
      Grimmer, Justin  and
      Stewart, Brandon M.",
    editor = "Al-Onaizan, Yaser  and
      Bansal, Mohit  and
      Chen, Yun-Nung",
    booktitle = "Proceedings of the 2024 Conference on Empirical Methods in Natural Language Processing",
    month = nov,
    year = "2024",
    address = "Miami, Florida, USA",
    publisher = "Association for Computational Linguistics",
    url = "https://aclanthology.org/2024.emnlp-main.913/",
    doi = "10.18653/v1/2024.emnlp-main.913",
    pages = "16325--16342"
}

@inproceedings{li-etal-2025-adaptive,
    title = "Adaptive Tool Use in Large Language Models with Meta-Cognition Trigger",
    author = "Li, Wenjun  and
      Li, Dexun  and
      Dong, Kuicai  and
      Zhang, Cong  and
      Zhang, Hao  and
      Liu, Weiwen  and
      Wang, Yasheng  and
      Tang, Ruiming  and
      Liu, Yong",
    editor = "Che, Wanxiang  and
      Nabende, Joyce  and
      Shutova, Ekaterina  and
      Pilehvar, Mohammad Taher",
    booktitle = "Proceedings of the 63rd Annual Meeting of the Association for Computational Linguistics (Volume 1: Long Papers)",
    month = jul,
    year = "2025",
    address = "Vienna, Austria",
    publisher = "Association for Computational Linguistics",
    url = "https://aclanthology.org/2025.acl-long.655/",
    doi = "10.18653/v1/2025.acl-long.655",
    pages = "13346--13370",
    ISBN = "979-8-89176-251-0"
}

@inproceedings{wang-etal-2025-meta,
    title = "Meta-Reflection: A Feedback-Free Reflection Learning Framework",
    author = "Wang, Yaoke  and
      Zhu, Yun  and
      XintongBao, XintongBao  and
      Zhang, Wenqiao  and
      Dai, Suyang  and
      Chen, Kehan  and
      Li, Wenqiang  and
      Huang, Gang  and
      Tang, Siliang  and
      Zhuang, Yueting",
    editor = "Che, Wanxiang  and
      Nabende, Joyce  and
      Shutova, Ekaterina  and
      Pilehvar, Mohammad Taher",
    booktitle = "Proceedings of the 63rd Annual Meeting of the Association for Computational Linguistics (Volume 1: Long Papers)",
    month = jul,
    year = "2025",
    address = "Vienna, Austria",
    publisher = "Association for Computational Linguistics",
    url = "https://aclanthology.org/2025.acl-long.201/",
    doi = "10.18653/v1/2025.acl-long.201",
    pages = "3958--3976",
    ISBN = "979-8-89176-251-0"
}

@inproceedings{
wan2025rema,
title={Re{MA}: Learning to Meta-Think for {LLM}s with Multi-agent Reinforcement Learning},
author={Ziyu Wan and Yunxiang LI and Xiaoyu Wen and Yan Song and Hanjing Wang and Linyi Yang and Mark Schmidt and Jun Wang and Weinan Zhang and Shuyue Hu and Ying Wen},
booktitle={The Thirty-ninth Annual Conference on Neural Information Processing Systems},
year={2025},
url={https://openreview.net/forum?id=ur295YVtmt}
}

@inproceedings{gu2025metawriter,
  title={MetaWriter: Personalized Handwritten Text Recognition Using Meta-Learned Prompt Tuning},
  author={Gu, Wenhao and Gu, Li and Suen, Chingyee Yee and Wang, Yang},
  booktitle={Proceedings of the Computer Vision and Pattern Recognition Conference},
  pages={23494--23504},
  year={2025}
}

@article{gong2025tuning,
  title={Tuning llm-based code optimization via meta-prompting: An industrial perspective},
  author={Gong, Jingzhi and Giavrimis, Rafail and Brookes, Paul and Voskanyan, Vardan and Wu, Fan and Ashiga, Mari and Truscott, Matthew and Basios, Mike and Kanthan, Leslie and Xu, Jie and others},
  journal={arXiv preprint arXiv:2508.01443},
  year={2025}
}

@inproceedings{wang-etal-2024-meta,
    title = "Meta-Reasoning: Semantics-Symbol Deconstruction for Large Language Models",
    author = "Wang, Yiming  and
      Zhang, Zhuosheng  and
      Zhang, Pei  and
      Yang, Baosong  and
      Wang, Rui",
    editor = "Ku, Lun-Wei  and
      Martins, Andre  and
      Srikumar, Vivek",
    booktitle = "Findings of the Association for Computational Linguistics: ACL 2024",
    month = aug,
    year = "2024",
    address = "Bangkok, Thailand",
    publisher = "Association for Computational Linguistics",
    url = "https://aclanthology.org/2024.findings-acl.34/",
    doi = "10.18653/v1/2024.findings-acl.34",
    pages = "622--643"
}

@inproceedings{wu-etal-2025-meta,
    title = "Meta-Rewarding Language Models: Self-Improving Alignment with {LLM}-as-a-Meta-Judge",
    author = "Wu, Tianhao  and
      Yuan, Weizhe  and
      Golovneva, Olga  and
      Xu, Jing  and
      Tian, Yuandong  and
      Jiao, Jiantao  and
      Weston, Jason E  and
      Sukhbaatar, Sainbayar",
    editor = "Christodoulopoulos, Christos  and
      Chakraborty, Tanmoy  and
      Rose, Carolyn  and
      Peng, Violet",
    booktitle = "Proceedings of the 2025 Conference on Empirical Methods in Natural Language Processing",
    month = nov,
    year = "2025",
    address = "Suzhou, China",
    publisher = "Association for Computational Linguistics",
    url = "https://aclanthology.org/2025.emnlp-main.583/",
    doi = "10.18653/v1/2025.emnlp-main.583",
    pages = "11548--11565",
    ISBN = "979-8-89176-332-6"
}

@inproceedings{he-etal-2025-gcml,
    title = "{GCML}: Gradient Coherence Guided Meta-Learning for Cross-Domain Emerging Topic Rumor Detection",
    author = "He, Zejiang  and
      Huang, Jingyuan  and
      Lu, Menglong  and
      Huang, Zhen  and
      Liu, Shanshan  and
      Tian, Zhiliang  and
      Li, Dongsheng",
    editor = "Christodoulopoulos, Christos  and
      Chakraborty, Tanmoy  and
      Rose, Carolyn  and
      Peng, Violet",
    booktitle = "Proceedings of the 2025 Conference on Empirical Methods in Natural Language Processing",
    month = nov,
    year = "2025",
    address = "Suzhou, China",
    publisher = "Association for Computational Linguistics",
    url = "https://aclanthology.org/2025.emnlp-main.968/",
    doi = "10.18653/v1/2025.emnlp-main.968",
    pages = "19159--19173",
    ISBN = "979-8-89176-332-6"
}

@inproceedings{qin-etal-2025-r,
    title = "{R}-{CHAR}: A Metacognition-Driven Framework for Role-Playing in Large Language Models",
    author = "Qin, Haiming  and
      Zhang, Jiwei  and
      Zhang, Wei  and
      Lu, KeZhong  and
      Zhou, Mingyang  and
      Liao, Hao  and
      Mao, Rui",
    editor = "Christodoulopoulos, Christos  and
      Chakraborty, Tanmoy  and
      Rose, Carolyn  and
      Peng, Violet",
    booktitle = "Proceedings of the 2025 Conference on Empirical Methods in Natural Language Processing",
    month = nov,
    year = "2025",
    address = "Suzhou, China",
    publisher = "Association for Computational Linguistics",
    url = "https://aclanthology.org/2025.emnlp-main.1372/",
    doi = "10.18653/v1/2025.emnlp-main.1372",
    pages = "26984--27002",
    ISBN = "979-8-89176-332-6"
}

@inproceedings{xiong-etal-2025-mpo,
    title = "{MPO}: Boosting {LLM} Agents with Meta Plan Optimization",
    author = "Xiong, Weimin  and
      Song, Yifan  and
      Dong, Qingxiu  and
      Zhao, Bingchan  and
      Song, Feifan  and
      XWang  and
      Li, Sujian",
    editor = "Christodoulopoulos, Christos  and
      Chakraborty, Tanmoy  and
      Rose, Carolyn  and
      Peng, Violet",
    booktitle = "Findings of the Association for Computational Linguistics: EMNLP 2025",
    month = nov,
    year = "2025",
    address = "Suzhou, China",
    publisher = "Association for Computational Linguistics",
    url = "https://aclanthology.org/2025.findings-emnlp.210/",
    doi = "10.18653/v1/2025.findings-emnlp.210",
    pages = "3914--3935",
    ISBN = "979-8-89176-335-7"
}

@inproceedings{zhang-etal-2025-remembering,
    title = "From Remembering to Metacognition: Do Existing Benchmarks Accurately Evaluate {LLM}s?",
    author = "Zhang, Geng  and
      Ying, Yizhou  and
      Jiang, Sihang  and
      Liang, Jiaqing  and
      Yue, Guanglei  and
      Fu, Yifei  and
      Hu, Hailin  and
      Xiao, Yanghua",
    editor = "Christodoulopoulos, Christos  and
      Chakraborty, Tanmoy  and
      Rose, Carolyn  and
      Peng, Violet",
    booktitle = "Findings of the Association for Computational Linguistics: EMNLP 2025",
    month = nov,
    year = "2025",
    address = "Suzhou, China",
    publisher = "Association for Computational Linguistics",
    url = "https://aclanthology.org/2025.findings-emnlp.724/",
    doi = "10.18653/v1/2025.findings-emnlp.724",
    pages = "13440--13457",
    ISBN = "979-8-89176-335-7"
}

@book{Metcalfe1994MetacognitionK,
  title     = {Metacognition: Knowing about Knowing},
  editor    = {Metcalfe, Janet and Shimamura, Arthur P.},
  publisher = {The MIT Press},
  address   = {Cambridge, MA},
  year      = {1994},
  note      = {A Bradford Book}
}

@inproceedings{li-etal-2024-meta,
    title = "Meta-Cognitive Analysis: Evaluating Declarative and Procedural Knowledge in Datasets and Large Language Models",
    author = "Li, Zhuoqun  and
      Lin, Hongyu  and
      Lu, Yaojie  and
      Xiang, Hao  and
      Han, Xianpei  and
      Sun, Le",
    editor = "Calzolari, Nicoletta  and
      Kan, Min-Yen  and
      Hoste, Veronique  and
      Lenci, Alessandro  and
      Sakti, Sakriani  and
      Xue, Nianwen",
    booktitle = "Proceedings of the 2024 Joint International Conference on Computational Linguistics, Language Resources and Evaluation (LREC-COLING 2024)",
    month = may,
    year = "2024",
    address = "Torino, Italia",
    publisher = "ELRA and ICCL",
    url = "https://aclanthology.org/2024.lrec-main.980/",
    pages = "11222--11228"
}

@misc{fan2025metacognitiveknowledgeeditingmultimodal,
      title={Towards Meta-Cognitive Knowledge Editing for Multimodal LLMs}, 
      author={Zhaoyu Fan and Kaihang Pan and Mingze Zhou and Bosheng Qin and Juncheng Li and Shengyu Zhang and Wenqiao Zhang and Siliang Tang and Fei Wu and Yueting Zhuang},
      year={2025},
      eprint={2509.05714},
      archivePrefix={arXiv},
      primaryClass={cs.AI},
      url={https://arxiv.org/abs/2509.05714}, 
}

@misc{ha2025ahamomentscontrollablethinking,
      title={From "Aha Moments" to Controllable Thinking: Toward Meta-Cognitive Reasoning in Large Reasoning Models via Decoupled Reasoning and Control}, 
      author={Rui Ha and Chaozhuo Li and Rui Pu and Sen Su},
      year={2025},
      eprint={2508.04460},
      archivePrefix={arXiv},
      primaryClass={cs.AI},
      url={https://arxiv.org/abs/2508.04460}, 
}

@misc{li2024metacognitiveanalysisevaluatingdeclarative,
      title={Meta-Cognitive Analysis: Evaluating Declarative and Procedural Knowledge in Datasets and Large Language Models}, 
      author={Zhuoqun Li and Hongyu Lin and Yaojie Lu and Hao Xiang and Xianpei Han and Le Sun},
      year={2024},
      eprint={2403.09750},
      archivePrefix={arXiv},
      primaryClass={cs.CL},
      url={https://arxiv.org/abs/2403.09750}, 
}

@inproceedings{
Xiao2025promptmii,
title={Prompt-{MII}: Meta-Learning Instruction Induction for {LLM}s},
author={Emily Xiao and Yixiao Zeng and Ada Chen and Chin-Jou Li and Amanda Bertsch and Graham Neubig},
booktitle={Submitted to The Fourteenth International Conference on Learning Representations},
year={2025},
url={https://openreview.net/forum?id=zD9fjEj4Oz},
note={under review}
}

@misc{sun2025metaragmetaanalysisinspiredevidencererankingmethod,
      title={META-RAG: Meta-Analysis-Inspired Evidence-Re-Ranking Method for Retrieval-Augmented Generation in Evidence-Based Medicine}, 
      author={Mengzhou Sun and Sendong Zhao and Jianyu Chen and Haochun Wang and Bing Qin},
      year={2025},
      eprint={2510.24003},
      archivePrefix={arXiv},
      primaryClass={cs.CL},
      url={https://arxiv.org/abs/2510.24003}, 
}

@inproceedings{wang-zhao-2024-metacognitive,
    title = "Metacognitive Prompting Improves Understanding in Large Language Models",
    author = "Wang, Yuqing  and
      Zhao, Yun",
    editor = "Duh, Kevin  and
      Gomez, Helena  and
      Bethard, Steven",
    booktitle = "Proceedings of the 2024 Conference of the North American Chapter of the Association for Computational Linguistics: Human Language Technologies (Volume 1: Long Papers)",
    month = jun,
    year = "2024",
    address = "Mexico City, Mexico",
    publisher = "Association for Computational Linguistics",
    url = "https://aclanthology.org/2024.naacl-long.106/",
    doi = "10.18653/v1/2024.naacl-long.106",
    pages = "1914--1926"
}

@misc{khan2025literalliberalmetapromptingframework,
      title={From Literal to Liberal: A Meta-Prompting Framework for Eliciting Human-Aligned Exception Handling in Large Language Models}, 
      author={Imran Khan},
      year={2025},
      eprint={2510.12864},
      archivePrefix={arXiv},
      primaryClass={cs.AI},
      url={https://arxiv.org/abs/2510.12864}, 
}

@misc{wang2025mralignmetareasoninginformedfactuality,
      title={MR-Align: Meta-Reasoning Informed Factuality Alignment for Large Reasoning Models}, 
      author={Xinming Wang and Jian Xu and Bin Yu and Sheng Lian and Hongzhu Yi and Yi Chen and Yingjian Zhu and Boran Wang and Hongming Yang and Han Hu and Xu-Yao Zhang and Cheng-Lin Liu},
      year={2025},
      eprint={2510.24794},
      archivePrefix={arXiv},
      primaryClass={cs.CL},
      url={https://arxiv.org/abs/2510.24794}, 
}

@misc{LeCun2022APT,
  title        = {A Path Towards Autonomous Machine Intelligence},
  author       = {LeCun, Yann},
  year         = {2022},
  note         = {Version 0.9.2, 2022-06-27},
  howpublished = {\url{https://api.semanticscholar.org/CorpusID:251881108}}
}

@inproceedings{jin-etal-2024-persuading,
    title = "Persuading across Diverse Domains: a Dataset and Persuasion Large Language Model",
    author = "Jin, Chuhao  and
      Ren, Kening  and
      Kong, Lingzhen  and
      Wang, Xiting  and
      Song, Ruihua  and
      Chen, Huan",
    editor = "Ku, Lun-Wei  and
      Martins, Andre  and
      Srikumar, Vivek",
    booktitle = "Proceedings of the 62nd Annual Meeting of the Association for Computational Linguistics (Volume 1: Long Papers)",
    month = aug,
    year = "2024",
    address = "Bangkok, Thailand",
    publisher = "Association for Computational Linguistics",
    url = "https://aclanthology.org/2024.acl-long.92/",
    doi = "10.18653/v1/2024.acl-long.92",
    pages = "1678--1706"
}

@book{cialdini2021influence,
  author    = {Robert B. Cialdini},
  title     = {Influence, New and Expanded: The Psychology of Persuasion},
  year      = {2021},
  publisher = {Harper Business}
}

@article{cohen1968weighted,
  title        = {Weighted kappa: Nominal scale agreement with provision for scaled disagreement or partial credit},
  author       = {Cohen, Jacob},
  journal      = {Psychological Bulletin},
  volume       = {70},
  number       = {4},
  pages        = {213--220},
  year         = {1968},
  publisher    = {American Psychological Association}
}

\appendix

\begin{figure*}[t]
    \centering
    \begin{subfigure}[t]{0.49\textwidth}
        \centering
        \includegraphics[width=\linewidth]{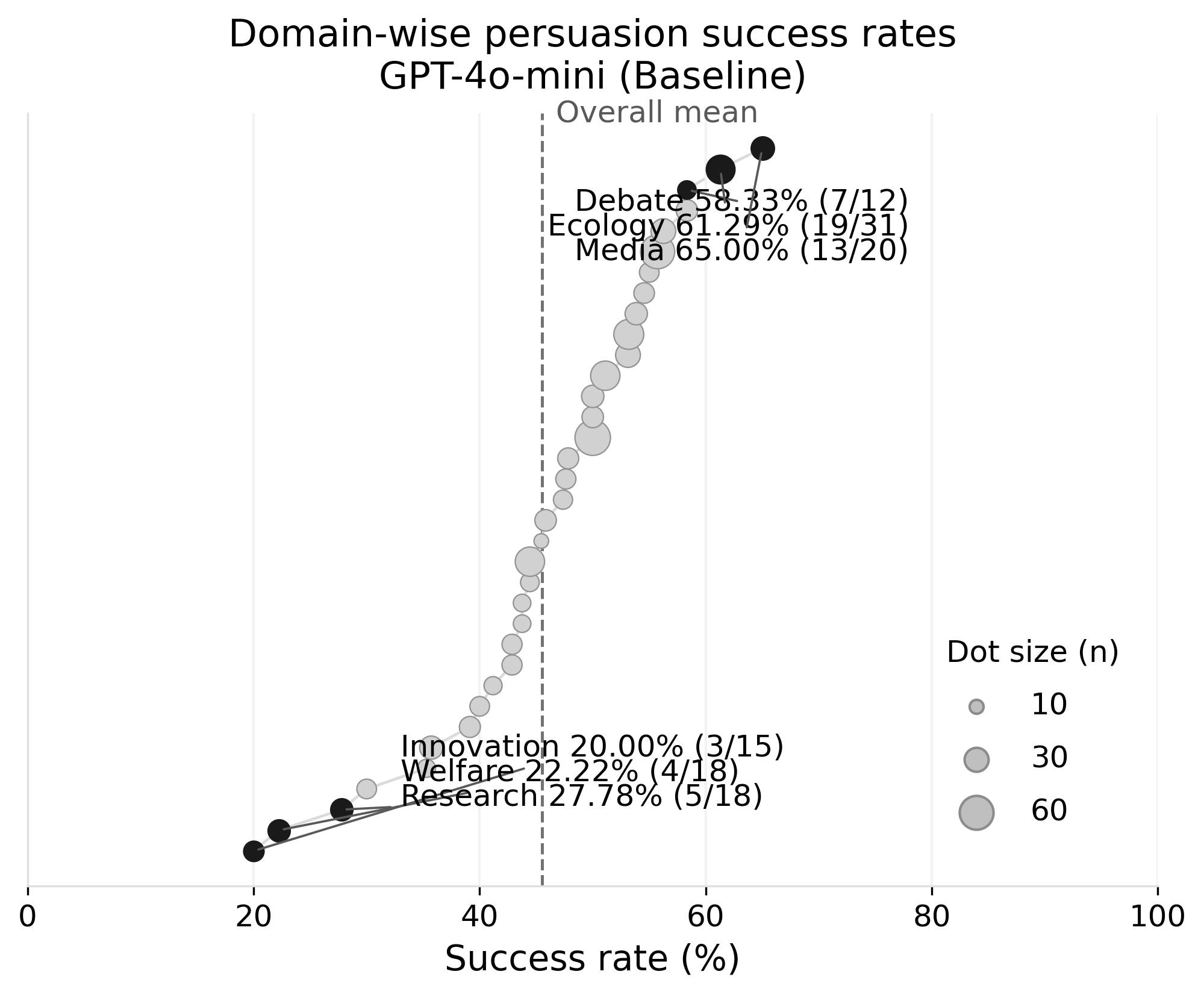}
        \caption{GPT-4o-mini (Baseline).}
        \label{fig:domain-success-4omini-base}
    \end{subfigure}
    \hfill
    \begin{subfigure}[t]{0.49\textwidth}
        \centering
        \includegraphics[width=\linewidth]{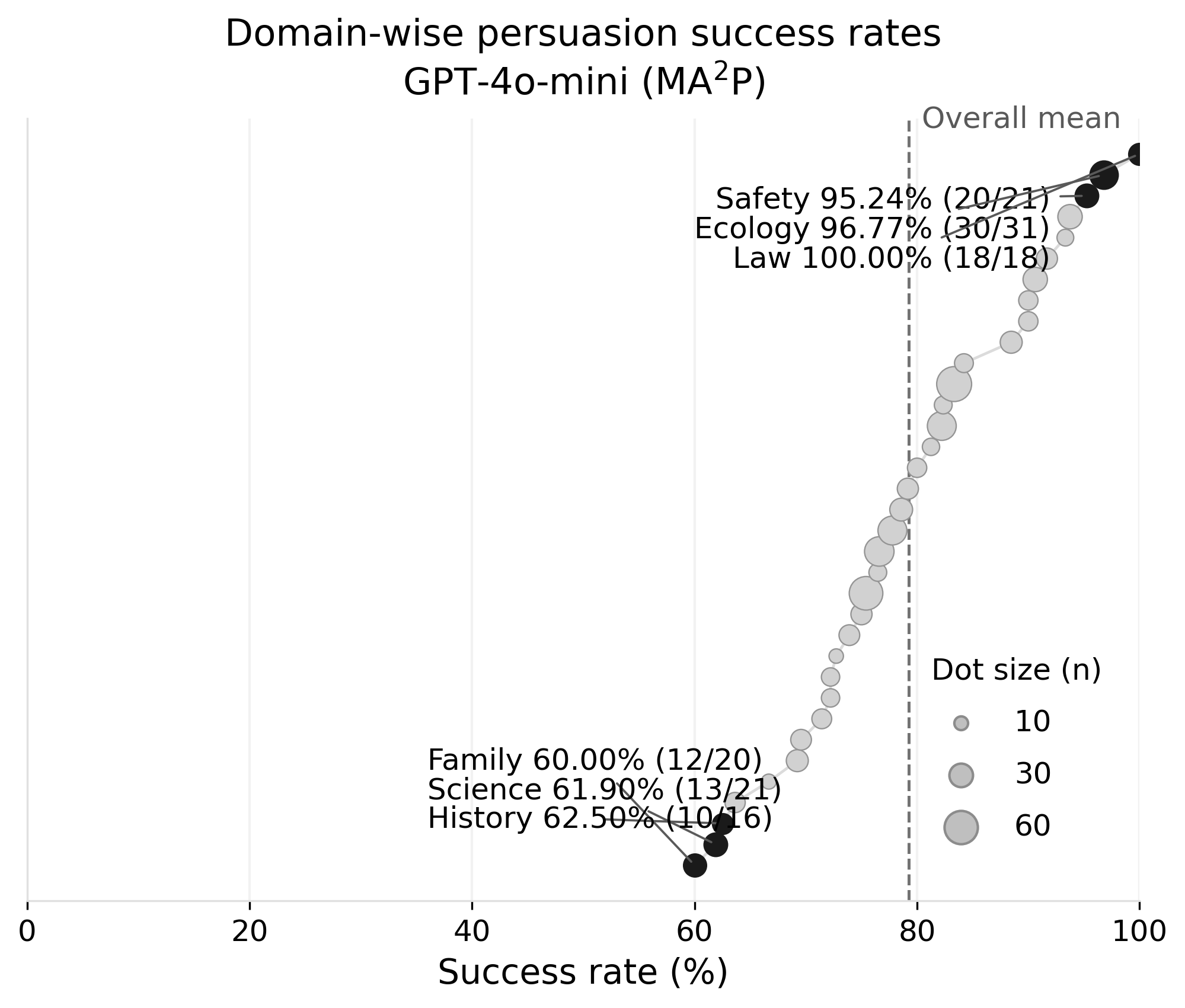}
        \caption{GPT-4o-mini (MA$^{2}$P).}
        \label{fig:domain-success-4omini-ma2p}
    \end{subfigure}
    \caption{Domain-wise persuasion success rates for GPT-4o-mini across domains.}
    \label{fig:domain-success-4omini}
    \vspace{-0.5cm}
\end{figure*}

\begin{figure*}[t]
    \centering
    \begin{subfigure}[t]{0.49\textwidth}
        \centering
        \includegraphics[width=\linewidth]{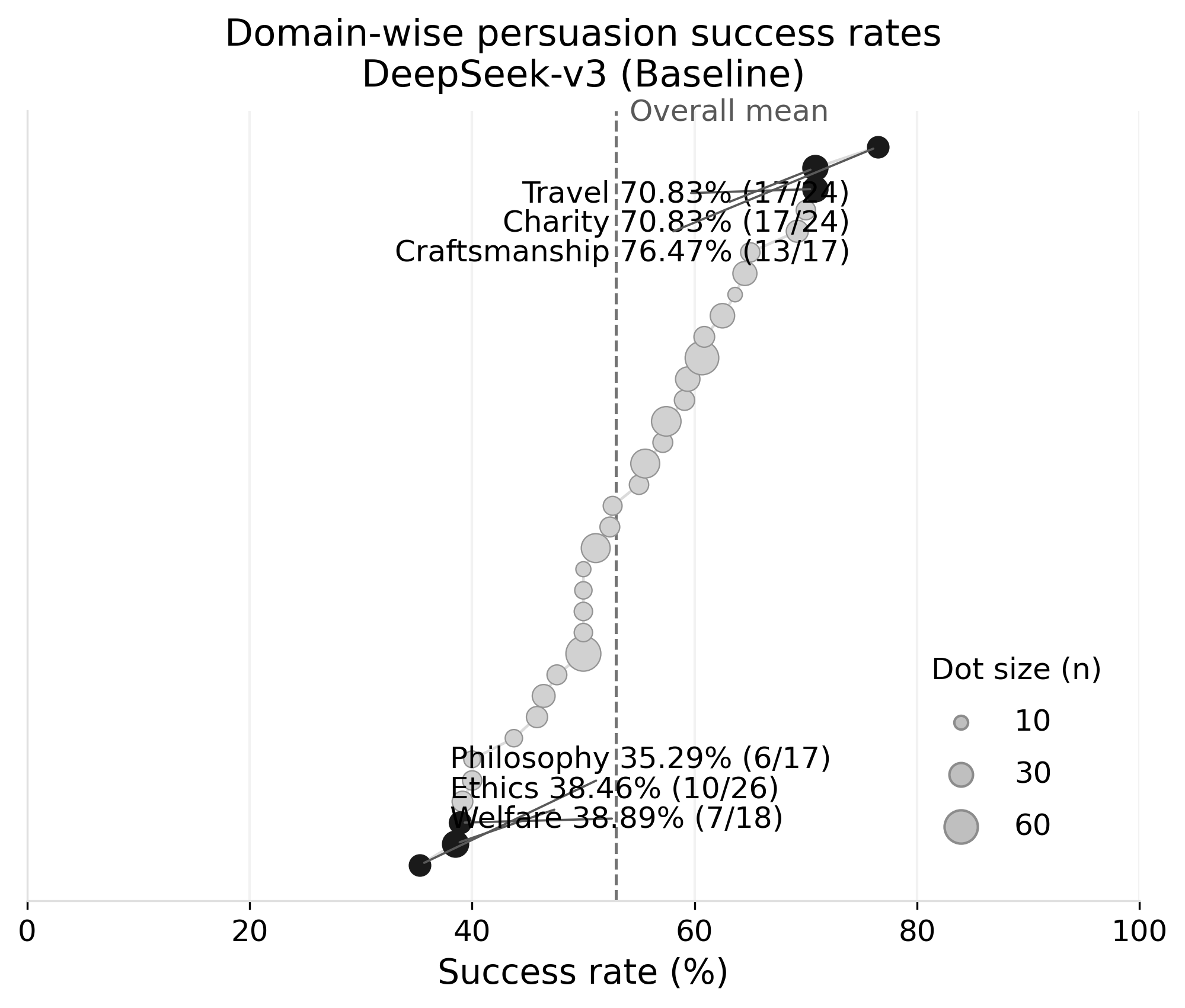}
        \caption{Deepseek-v3 (Baseline).}
        \label{fig:domain-success-dsv3-base}
    \end{subfigure}
    \hfill
    \begin{subfigure}[t]{0.49\textwidth}
        \centering
        \includegraphics[width=\linewidth]{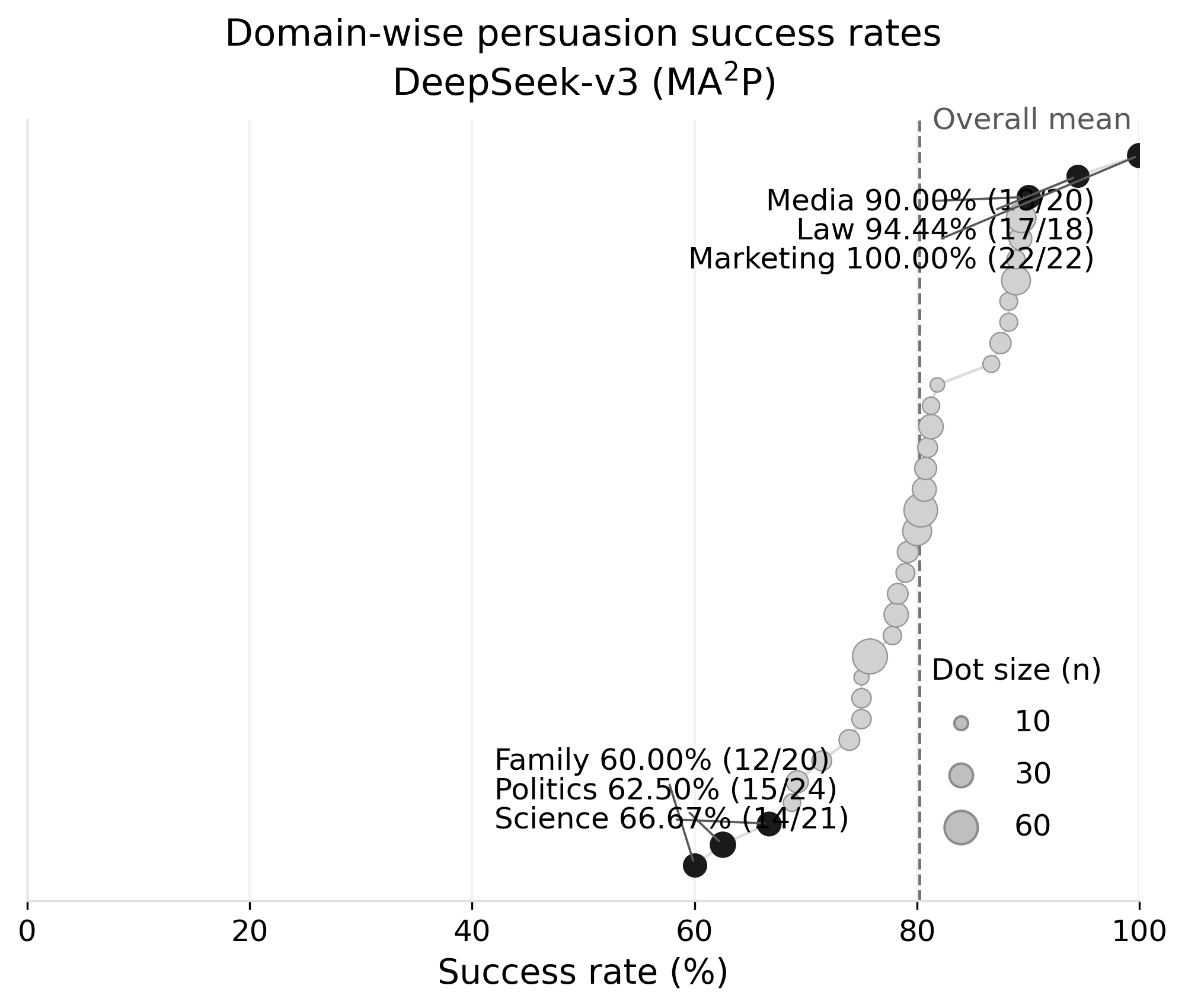}
        \caption{Deepseek-v3 (MA$^{2}$P).}
        \label{fig:domain-success-dsv3-ma2p}
    \end{subfigure}
    \caption{Domain-wise persuasion success rates for deepseek-v3 across domains. Each dot corresponds to one domain; dot size indicates the number of test instances ($n$). The vertical dashed line marks the overall mean success rate. We annotate the top-3 and bottom-3 domains to highlight best- and worst-performing cases.}
    \label{fig:domain-success-dsv3}
\end{figure*}
\FloatBarrier

\section{Other experiments}
\subsection{Domain Success Rate Visualization}
The differences in success rates across domains are illustrated in Fig.~\ref{fig:domain-success-4omini} and Fig.~\ref{fig:domain-success-dsv3}. 
Each dot corresponds to one domain, with the dot size indicating the number of test instances ($n$) in that domain; the vertical dashed line marks the overall mean success rate. 
For both \texttt{gpt-4o-mini} and \texttt{deepseek-v3}, MA\textsuperscript{2}P shifts the domain-wise distribution to the right, indicating higher success rates across most domains, and visibly reduces extreme low-performing cases (annotated bottom-3 domains). 
Meanwhile, the performance becomes more concentrated around the mean: the gap between the best- and worst-performing domains narrows, which is consistent with the reduced dispersion metrics (\textbf{Range} and \textbf{SD}) reported in Table~\ref{tab:main_results}. 
Overall, these visualizations provide qualitative evidence that MA\textsuperscript{2}P improves cross-domain robustness in addition to increasing average persuasion success.

\begin{figure*}[t]
    \centering
    \begin{subfigure}[t]{0.49\textwidth}
        \centering
        \includegraphics[width=\linewidth]{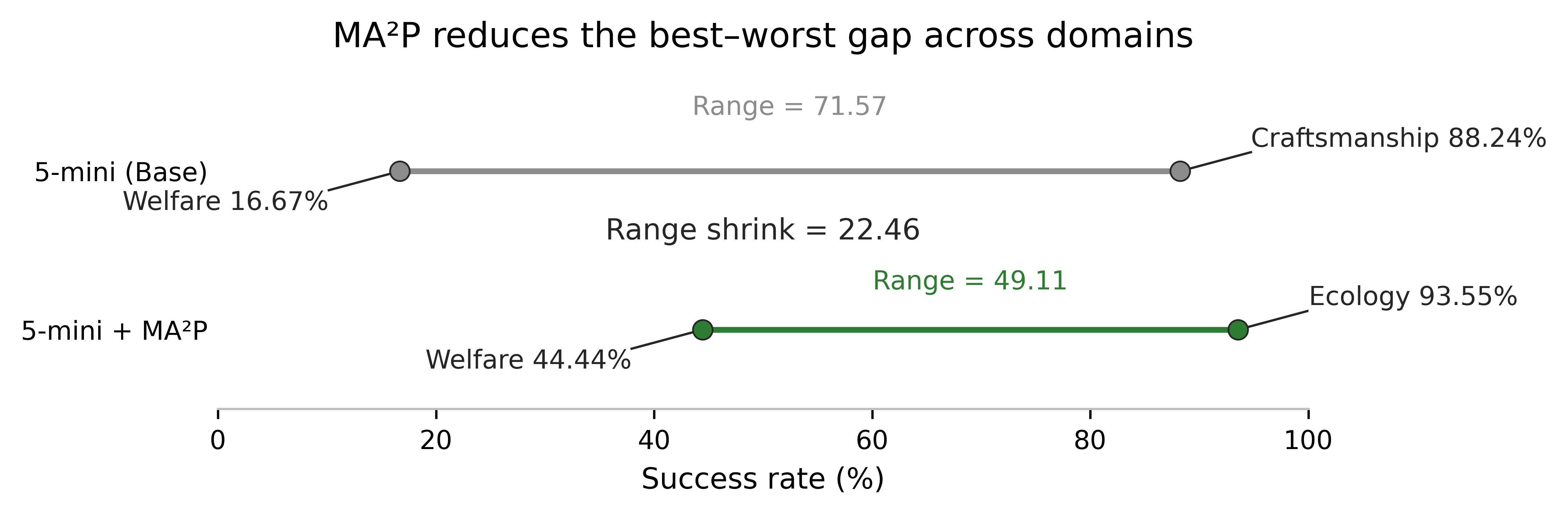}
        \caption{MA\textsuperscript{2}P reduces the best--worst gap (Range) on \texttt{gpt-5-mini}.}
        \label{fig:robustness_range}
    \end{subfigure}
    \hfill
    \begin{subfigure}[t]{0.49\textwidth}
        \centering
        \includegraphics[width=\linewidth]{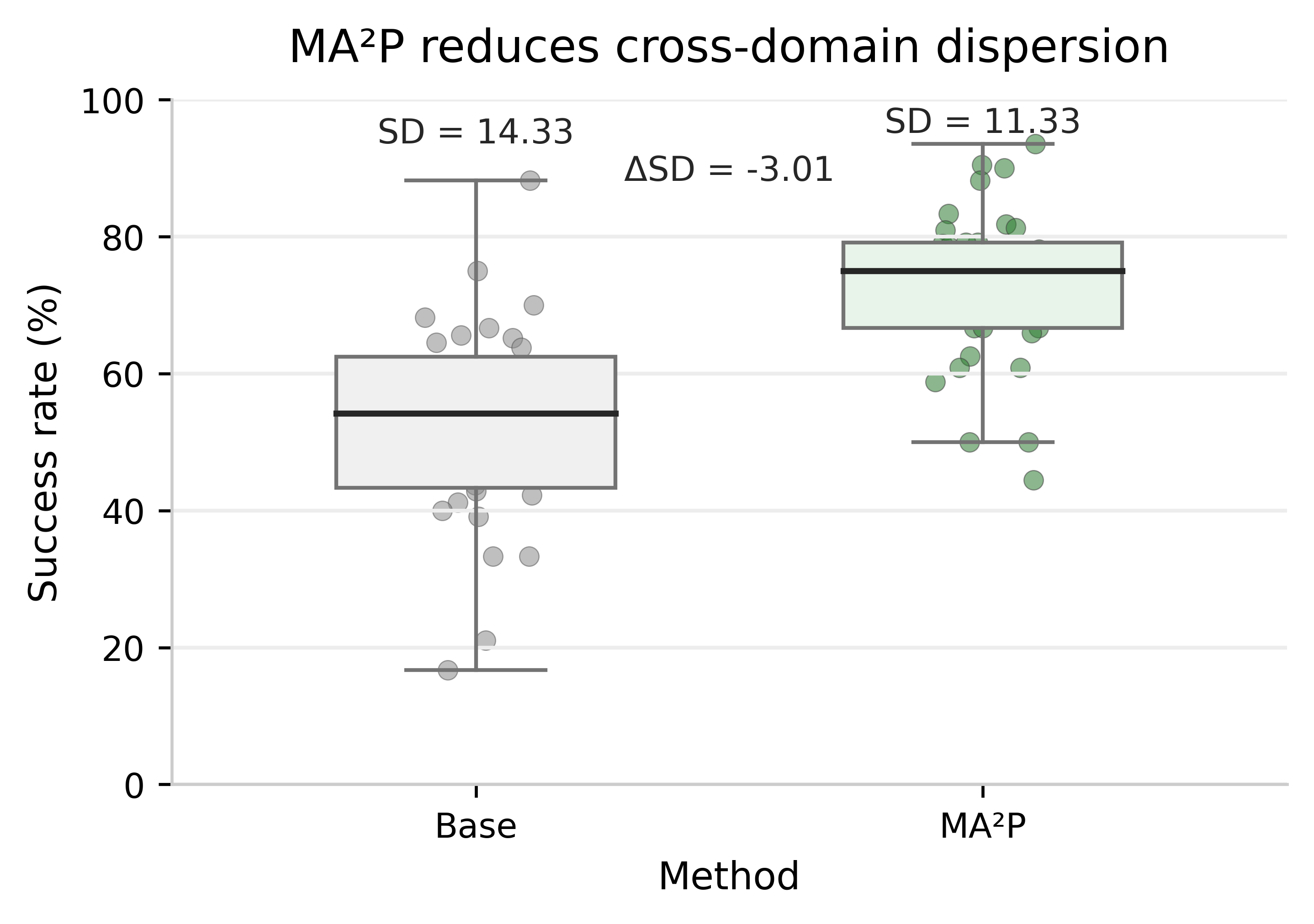}
        \caption{MA\textsuperscript{2}P reduces cross-domain dispersion (SD) on \texttt{gpt-5-mini}.}
        \label{fig:robustness_sd}
    \end{subfigure}
    \caption{Cross-domain robustness visualization on \texttt{gpt-5-mini}.}
    \label{fig:robustness_gpt5mini}
    \vspace{-0.2cm}
\end{figure*}

\begin{figure*}[t]
    \centering
    \begin{subfigure}[t]{0.49\textwidth}
        \centering
        \includegraphics[width=\linewidth]{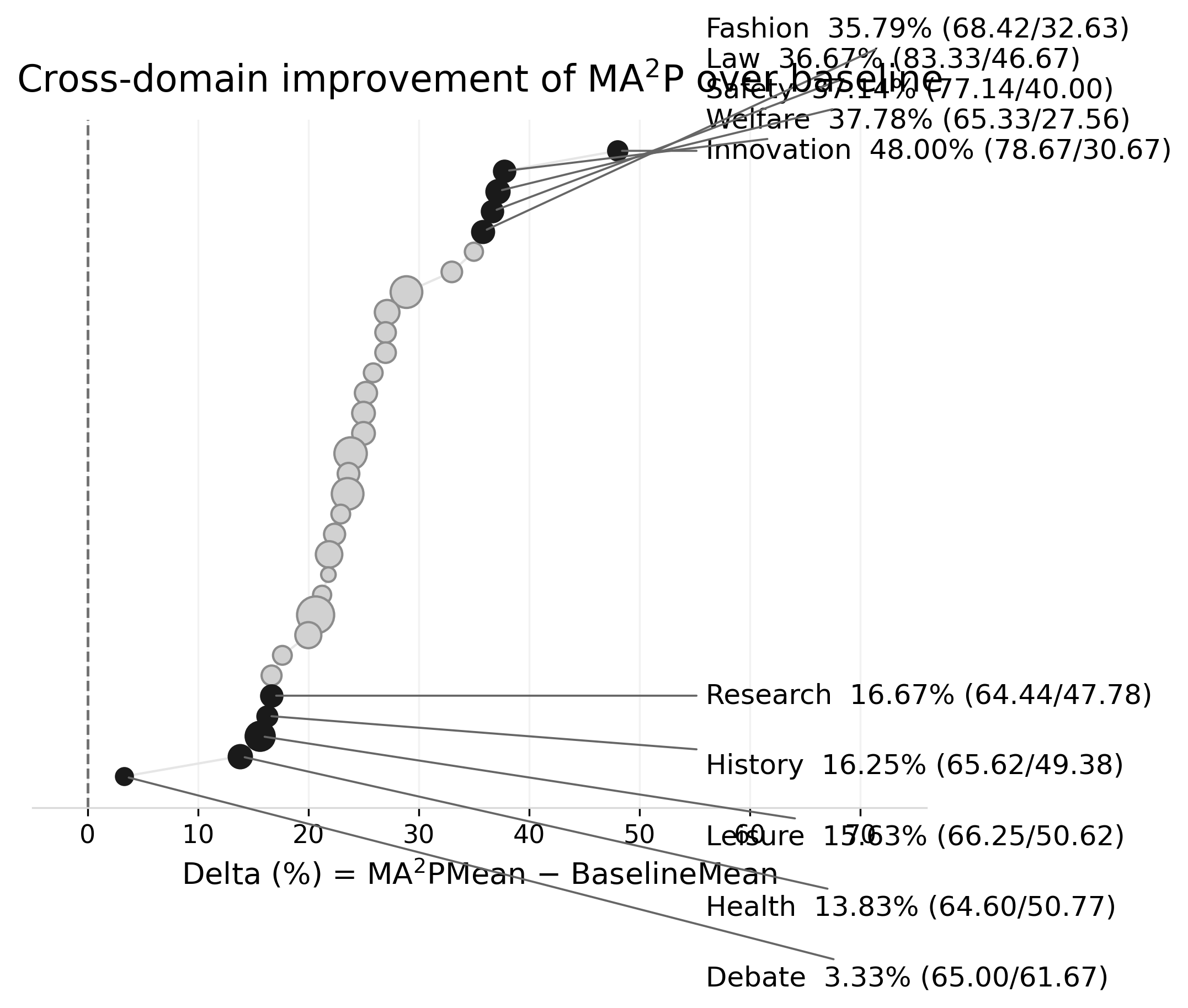}
        \caption{Cross-domain improvement of MA$^{2}$P over baseline.}
        \label{fig:delta-map}
    \end{subfigure}
    \hfill
    \begin{subfigure}[t]{0.49\textwidth}
        \centering
        \includegraphics[width=\linewidth]{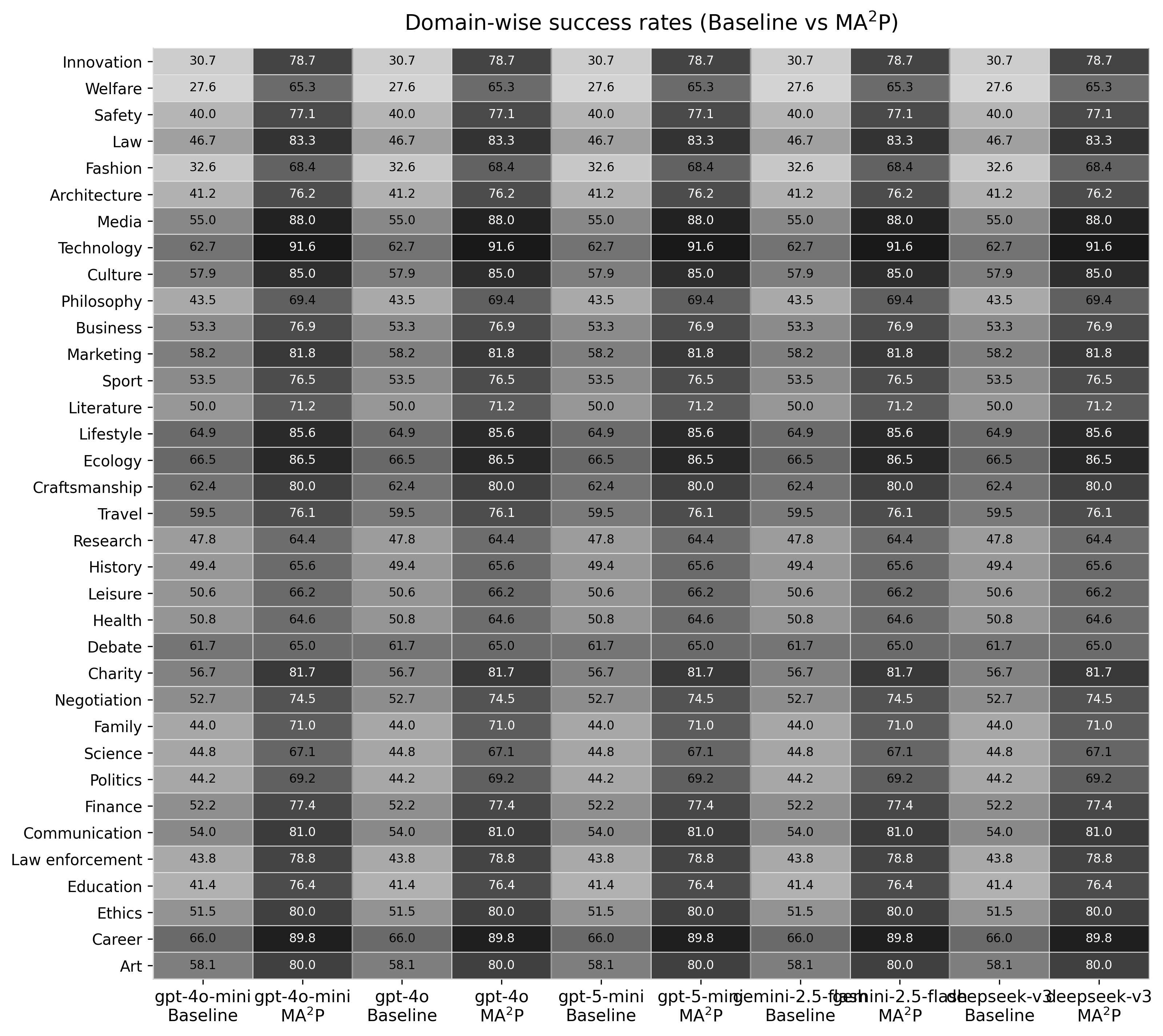}
        \caption{Domain-wise success rates (baseline vs.\ MA$^{2}$P) across five base models.}
        \label{fig:domain-heatmap}
    \end{subfigure}
    \caption{Additional visualizations of domain-wise success rates and cross-domain improvements.}
    \label{fig:other-vis}
    \vspace{-0.2cm}
\end{figure*}

\subsection{Cross-domain Robustness Visualization}
To make cross-domain dispersion more interpretable, we visualize domain-wise success rates for \texttt{gpt-5-mini} and its MA\textsuperscript{2}P-augmented counterpart.
As shown in Fig.~\ref{fig:robustness_range}, the best--worst gap shrinks from 71.57 percentage points to 49.11 percentage points.
Meanwhile, Fig.~\ref{fig:robustness_sd} shows that the standard deviation across the 35 domains drops from 14.33 to 11.33 percentage points.
Together, these results indicate that MA\textsuperscript{2}P reduces cross-domain variance by lifting weaker domains and narrowing the performance spread.

\subsection{Other Visualization}
\label{app:other-vis}

Figure~\ref{fig:delta-map} and Figure~\ref{fig:domain-heatmap} provide additional domain-level evidence.
First, under the \emph{baseline} setting, some domains are consistently ``easy'' across all five base models (success rate $\ge 50\%$), including \textsc{Career}, \textsc{Lifestyle}, \textsc{Art}, \textsc{Charity}, and \textsc{Debate}; in contrast, \textsc{Welfare} and \textsc{Innovation} are consistently ``hard'' (success rate $\le 40\%$).
Second, under MA$^{2}$P, several domains become stably strong across all base models (success rate $\ge 70\%$), including \textsc{Safety}, \textsc{Literature}, \textsc{Ecology}, \textsc{Negotiation}, and \textsc{Art}.
Third, the largest average gains (mean improvement over five models, in percentage points) concentrate on domains that are difficult for the baseline, e.g., \textsc{Innovation} (+48.0pp) and \textsc{Welfare} (+37.8pp), while domains with limited headroom (already relatively easy under the baseline) show smaller improvements, e.g., \textsc{Debate} (+3.3pp).
We also observe a small number of domain$\times$model cases with slight regressions (e.g., \textsc{Debate}, and occasionally \textsc{Finance}/\textsc{Health}/\textsc{Research}), suggesting that the best interaction style in certain domains may be closer to direct rebuttal or conversational sparring, where additional constraints can be unnecessary.

\begin{table*}[t]
\centering
\caption{Comparison with current method}
\begin{tabular}{llccc}
\toprule
Model & Setting & Success ($\uparrow$) & Range ($\downarrow$) & SD ($\downarrow$) \\
\midrule
\multirow{3}{*}{gpt-4o-mini}
& \citep{furumai-etal-2024-zero}   & 0.21 & \underline{0.500} & \textbf{0.097} \\
& \citep{karande-etal-2024-persuasion} & \underline{0.43} & 0.640 & 0.114 \\
& MA\textsuperscript{2}P    & \textbf{0.79} & \textbf{0.400} & \underline{0.107} \\
\midrule
\multirow{3}{*}{gemini-2.5-flash}
& \citep{furumai-etal-2024-zero}   & 0.18 & \underline{0.450} & \textbf{0.093} \\
& \citep{karande-etal-2024-persuasion} & \underline{0.44} & 0.680 & 0.140 \\
& MA\textsuperscript{2}P    & \textbf{0.69} & \textbf{0.422} & \underline{0.106} \\
\midrule
\multirow{3}{*}{deepseek-v3}
& \citep{furumai-etal-2024-zero}   & 0.19 & \textbf{0.319} & \underline{0.091} \\
& \citep{karande-etal-2024-persuasion} & \underline{0.59} & 0.510 & 0.124 \\
& MA\textsuperscript{2}P    & \textbf{0.80} & \underline{0.400} & \textbf{0.088} \\
\bottomrule
\end{tabular}
\label{tab:cm_compare}
\end{table*}

Finally, note that domain-wise success rates should be interpreted with care: strong (or weak) performance in a domain may reflect not only the persuader's capability, but also the \emph{persuadee}'s susceptibility and the dataset's domain-specific scenario composition (i.e., some domains may be intrinsically easier for the simulated persuadee to accept).

\subsection{Comparison with current method}
Table~\ref{tab:cm_compare} reports an additional comparison with two recent prompt-based persuasion approaches that appear most compatible with our evaluation setting.
To the best of our knowledge, there is currently no existing method that fully matches our \emph{multi-domain} and \emph{belief--desire-grounded} persuasion setup; therefore, we select two representative methods and adapt them as faithfully as possible.
Since neither method releases code, we implement them by following the prompts and procedures described in their papers.

Overall, both adapted baselines perform poorly in our setting, and in some cases are even weaker than the corresponding base model.
We attribute this primarily to a task mismatch: for example, \citet{furumai-etal-2024-zero} is developed under domain-restricted scenarios (e.g., recommendation, donation, and health) and does not directly address broad multi-domain persuasion, making it difficult to generalize to the 35-domain benchmark used in this work.
Because this comparison is inherently limited by such incompatibilities, we place it in the appendix and treat it as supplementary evidence rather than a main claim.

Despite these caveats, MA\textsuperscript{2}P achieves substantially higher persuasion success across all three base models in Table~\ref{tab:cm_compare}, indicating a clear advantage in goal attainment.
For dispersion metrics, \textbf{Range} and \textbf{SD} for the two prior methods can appear smaller largely because their overall success rates are low (making domain-wise rates more compressed). 
Even under this effect, MA\textsuperscript{2}P remains competitive on \textbf{Range}/\textbf{SD} (ranking within the top two in most cases) while delivering the strongest \textbf{Success} by a wide margin.

\section{Supplementary Notes on the Evaluation Procedure}
\label{app:procedure_notes}

We follow the official test split provided by \citet{zhang2025persuasiondoubleblindmultidomaindialogue}, where the authors designate 525 instances as the test set.

In early runs, the knowledge base (KB) can be empty. To avoid an unrealistic setup where the world model receives meta strategies without any prior experience, we initialize MA\textsuperscript{2}P with a \emph{no-KB} variant, i.e., the world-model prompt is not provided with any meta-strategy input.

We use 500 instances outside the 525-test split as an offline pool. Specifically, we use 200 instances as ``training data'' to populate an initial KB. After obtaining this initial KB, we perform KB updates on the remaining 300 instances using the same update procedure described in the main paper.

During evaluation on the 525-test split, we keep the KB fixed and do not perform online updates.

\section{Autonomous Intelligent Agent Prompt Templates}
\lstset{style=promptstyle}

\paragraph{Placeholders.}
Curly braces \texttt{\{\}} indicate runtime slots to be filled with the corresponding inputs
(e.g., conversation record, background, goal, strategies, preventive/generative, etc.).

\paragraph{First Round.}
Because the persuader speaks first in the initial turn—before receiving any reply from the persuadee—it must initiate the topic and set the conversational direction, which requires different strategies and phrasing than later turns; therefore, we use separate prompts for both the world model and the persuader in the first round.

\subsection{World Model (First Round Strategy)}
\begin{lstlisting}[caption={World model prompt for first-round strategies},label={app:prompt-wm-first}]
You are a persuasion expert who can devise persuasion strategies based on the given persuasion goals and information.

You will receive:
1. Background: {}
2. Persuasion goal: {}

Your role and task:
- Think from the perspective of the person being persuaded.
- Use the background and persuasion goal to infer what strategies are most likely to be effective.
- Focus on strategies suitable for the first round of persuasion.

Strategy requirements:
- You need to develop fewer than 5 persuasion strategies.
- Each strategy should be a (strategy name -> specific strategy) pair.
- The specific strategy descriptions should be concise and clearly instruct how to persuade.
- All strategies should be closely aligned with the given persuasion goal.

Output format (very important):
- Do not provide any explanations or notes outside the JSON.
- Output a single JSON object in the following format:
"strategy": {
          "": "",
          "": "",
          "": "",
        ...... 
        }
\end{lstlisting}

\subsection{World Model}
\begin{lstlisting}[caption={World model prompt for refining high-level strategy},label={app:prompt-wm}]
You are a persuasion expert who can devise persuasion strategies based on the given persuasion goals and information.

You will receive:
1. Dialogue record: {}
2. Background: {}
3. Persuasion goal: {}
4. The preventive{} and generative{} of the person to be persuaded.
5. High Level Strategy: {}.

Your role and task:
- Think from the perspective of the person being persuaded.
- Fully utilize all the given information: dialogue, background, goal, preventive, generative, and the high-level strategy.
- Refine the high-level strategy into concrete, implementable persuasion strategies.

Strategy requirements:
- You need to develop exactly 5 persuasion strategies.
- Each strategy should be a (strategy name -> specific strategy) pair.
- The specific strategy descriptions should be concise, clear, and operational.
- All strategies should be consistent with and derived from the given high-level strategy and persuasion goal.

Output format (very important):
- Do not provide any explanations or notes outside the JSON.
- Output a single JSON object in the following format:
"strategy": {
          "": "",
          "": "",
        ...... 
        }
\end{lstlisting}

\subsection{Persuader (First Round)}
\begin{lstlisting}[caption={Persuader prompt for first sentence},label={app:prompt-persuader-first}]
You are an experienced persuader skilled in various scenarios.

You will receive:
1. Background: {}
2. Persuasion goal: {}
3. Domains involved in the persuasion task: {}
4. Persuasion strategies suggested for the first round of dialogue{}

Your role and task:
- Think from the perspective of the person being persuaded, but speak as the persuader.
- Use the given background, persuasion goal, domains, and suggested first-round strategies.
- Design a natural, engaging first sentence that can open the conversation and lead toward the persuasion goal.

Behavior guidelines:
- Persuasion may require some discussion, so your tone should be open and inviting.
- Your tone, emotions, and wording should:
  - Not make the persuadee feel uncomfortable or embarrassed.
  - Fit the context and domain of the persuasion task.

Output requirements:
- Based on the above requirements, say the first sentence of the conversation.
- The first sentence should be brief and within two sentences.
- The sentence must start with "persuader:".
\end{lstlisting}

\subsection{Persuader (Multi-turn)}
\begin{lstlisting}[caption={Persuader prompt for next-turn response},label={app:prompt-persuader}]
You are an experienced persuader who can adapt to various situations.

You will receive:
1. Dialogue record: {}
2. Background: {}
3. Persuasion goal: {}
4. Suggested persuasion strategies: {}
5. The preventive{} and generative{} of the person to be persuaded 

Your role and task:
- Think from the perspective of the person being persuaded, but respond as the persuader.
- Use the dialogue record, background, goal, suggested strategies, and preventive/generative information.
- Determine how to speak in the next turn so as to move closer to achieving the persuasion goal.

Behavior guidelines:
- Persuasion may involve multiple rounds of discussion.
- You should actively guide the other person's thinking in a natural and respectful way.
- Your response should be coherent with the existing dialogue and clearly oriented toward the persuasion goal.

Output requirements:
- Based on the conversation at hand, say your next sentence as the persuader.
- The sentence should be brief and within three sentences.
- The sentence must start with "persuader:".
\end{lstlisting}

\subsection{Perception (Preventive/Generative Inference)}
\begin{lstlisting}[caption={Prompt for inferring preventive/generative states},label={app:prompt-perception}]
You are a master of psychology who can analyze people's preventives and generatives from conversations.

You will receive:
1. Background: {}
2. Persuasion goal: {}
3. Dialogue record: {}

Your role and task:
- Use the background, goal, and dialogue to infer:
  - The preventive content, beliefs, and desires of the persuadee.
  - The generative content, beliefs, and desires of the persuadee.
- Here:
  - "Preventive" corresponds to what the persuadee should NOT do in relation to the persuasion goal.
  - "Generative" corresponds to what the persuadee SHOULD do in relation to the persuasion goal.
- If there is no information available for an item, use "none" for that field.

Output requirements:
- Do not provide any explanations, notes, or extra text outside the JSON.
- Present your inferences strictly in the following JSON format:
"preventive": {
          "content": "",
          "belief": "",
          "desire": ""
        },
"generative": {
          "content": "",
          "belief": "",
          "desire": ""
        }
\end{lstlisting}

\section{Persuadee Prompt}

\begin{lstlisting}[caption={Persuadee simulation prompt},label={app:prompt-persuadee}]
When reading the dialog, please put yourself in the position of the persuadee.

You will receive:
1. dialogue{}
2. background{}
3. preventive{}
4. generative{}
5. end_flag{}

Role and behavior setting:
- You are simulating the real reactions of a potential persuadee in this specific scenario.
- Your tone, emotions, reactions, and wording must match:
  - The personality of the persuadee's role.
  - The background and settings of this scenario.
- The persuadee does not need to be overly polite; natural, realistic responses are preferred.
- If both your belief and Desire are satisfied, you should accept and take the advice.

Conversation rules:
- You should respond as the persuadee, continuing the conversation naturally.
- If end_flag is marked as true:
  - You must clearly express your final attitude toward the persuader.
  - You must end the conversation with a declarative sentence.

Output requirements:
- Reply with the next sentence based on the current conversation.
- The reply should be brief and within two sentences.
- The reply must start with "persuadee:".
\end{lstlisting}

\section{Evaluation Prompt}
\subsection{Persuasion Success Judge}
\begin{lstlisting}[caption={Judge prompt for persuasion success},label={app:prompt-judge}]
You are an expert in judging whether persuasion has been successful based on the conversation record.

You will receive:
1. Conversation record: {}
2. Persuasion goal: {}

Judgment criteria:
- Consider the persuasion successful if the persuadee:
  - Shows clear interest in the persuasion goal, or
  - Expresses approval or agreement with the persuasion goal, or
  - Expresses a desire to learn more or move forward with the persuasion goal, or
  - If, when you put yourself in the position of the persuadee, you believe you have already been persuaded.

Output requirements:
- Do not provide any explanation, reasoning, or additional content.
- If the persuasion is successful, return "True".
- If the persuasion is unsuccessful, return "False".
\end{lstlisting}

\subsection{A/B Preference Evaluation}
\begin{lstlisting}[caption={A/B prompt for pairwise persuasiveness comparison},label={app:prompt-ab}]
You are evaluating two persuasion dialogues for the same scenario.

Context (use this to ground your evaluation):
- Background (scenario description): {background}
- Preventive information (the persuadee's current interests/concerns and why): {preventive}
- Generative information (the intended direction/goal and why): {generative}

Dialogue 1:
{dialogue_1} 

Dialogue 2:
{dialogue_2}

Task:
You must follow this procedure:

Step 1) Evaluate each dialogue independently (do NOT compare yet).
For EACH dialogue, think about how strong it is overall in this scenario and how well it performs on:
- Understanding and addressing what the persuadee truly cares about
- Responding effectively to the concerns raised
- Providing meaningful, specific, and well-developed reasoning
- Adapting to the complexity/high-stakes nature of the decision
Then form an overall impression of its persuasive strength (e.g., weak / moderate / strong / very strong) grounded in the scenario.

Step 2) Compare Dialogue 1 vs Dialogue 2 only after Step 1.
When comparing, focus on the magnitude of the difference in overall persuasive strength, not small stylistic differences.

Final decision (wrapped in ### at the very end):
Choose ONE of:
   ###1. More Persuasive: Dialogue 1###
   ###2. More Persuasive: Dialogue 2###
   ###3. Equally Persuasive: Both dialogues###

Decision policy (EXTREMELY IMPORTANT - ties should be common):
- Default to option 3.
- Choose 1 or 2 ONLY in the rare case where ONE dialogue is OBVIOUSLY and SUBSTANTIALLY better than the other.
  "OBVIOUSLY and SUBSTANTIALLY better" means:
  - It outperforms the other dialogue across MOST of the criteria (not just one), AND
  - The advantage is large enough that a reasonable reader would clearly notice it, AND
  - The better dialogue has no major weaknesses relative to the other.
- If the difference is moderate, mixed, subtle, uncertain, or mainly stylistic, choose 3.
- If each dialogue has different strengths, choose 3 unless one dialogue overwhelmingly dominates overall.

Rules:
- The final decision MUST be wrapped with ### and MUST match one of the three exact option strings above.
- Put the final ###...### decision at the very end of your output.
\end{lstlisting}

\subsection{Persuasiveness Score}
\begin{lstlisting}[caption={Prompt for persuasiveness scoring},label={app:prompt-score-persuasive}]
Persuasiveness:

You need to evaluate the Persuasive strength of the given response.
Please imagine that you are the persuadee in the conversation.

Persuasive means: The response should aim to change the persuadee's mind.

The score must be an integer from 1 to 10.
1 means the worst and 10 means the best.
You may use intermediate values such as 2, 3, 5, 7, or 9.

Scoring criteria:

1: When I saw this reply, I became even more determined not to follow the persuader's view.

4: When I saw this reply, I still firmly held my original opinion.

6: When I saw this reply, I felt neutral and had no clear reaction.

8: When I saw this reply, I was somewhat moved or persuaded.

10: When I saw this reply, I strongly agreed with the persuader and changed my mind.

Please strictly follow this output format:
Persuasive: <int>

You ONLY need to give a score.
Do NOT include any explanation.
Please use STRICT STANDARDS when giving a score.

--------------------
Here is the conversation context you should consider:

Background:
{}

Dialogue:
{}
\end{lstlisting}

\subsection{Logical-Coherence Score}
\begin{lstlisting}[caption={Prompt for logical-coherence scoring},label={app:prompt-score-logical}]
Logical-Coherence:

You need to evaluate the Logical-Coherence of the given response.
Please imagine that you are the persuadee in the conversation.

Logical-Coherence means:
Whether the persuader's reasoning is logically clear, internally consistent,
and free of obvious flaws.

The score must be an integer from 1 to 10.
1 means the worst and 10 means the best.
You may use intermediate values such as 2, 3, 5, 7, or 9.

Scoring criteria:

1: When I saw this reply, the logic was confusing or self-contradictory,
and the argument failed to make sense.

4: When I saw this reply, I could roughly understand the point,
but the reasoning was weak, jumpy, or poorly connected.

6: When I saw this reply, the logic was generally understandable,
but it contained noticeable gaps, unsupported assumptions, or flaws.

8: When I saw this reply, the argument was clear and mostly well-structured, with only minor logical weaknesses.

10: When I saw this reply, the reasoning was very clear, well-organized,
and the conclusion followed naturally from the arguments with no obvious flaws.

Please strictly follow this output format:
Logical-Coherence: <int>

You ONLY need to give a score.
Do NOT include any explanation.
Please use STRICT STANDARDS when giving a score.

--------------------
Here is the conversation context you should consider:

Background:
{}

Dialogue:
{}
\end{lstlisting}

\subsection{Helpfulness Score}
\begin{lstlisting}[caption={Prompt for helpfulness scoring},label={app:prompt-score-helpful}]
Helpfulness:

You need to evaluate the Helpfulness of the given response.
Please imagine that you are the persuadee in the conversation.

Helpfulness means:
Whether the response is useful, relevant, and helpful for addressing
the persuadee's question, concern, or situation.

The score must be an integer from 1 to 10.
1 means the worst and 10 means the best.
You may use intermediate values such as 2, 3, 5, 7, or 9.

Scoring criteria:

1: When I saw this reply, it was completely unhelpful and did not address my needs.

4: When I saw this reply, it was only loosely related to my situation and provided little practical help.

6: When I saw this reply, it offered some relevant information, but the help was limited or incomplete.

8: When I saw this reply, it addressed my concern well and was genuinely helpful for understanding or decision-making.

10: When I saw this reply, it was highly relevant, precise, and effectively helped resolve my problem or confusion.

Please strictly follow this output format:
Helpfulness: <int>

You ONLY need to give a score.
Do NOT include any explanation.
Please use STRICT STANDARDS when giving a score.

--------------------
Here is the conversation context you should consider:

Background:
{}

Dialogue:
{}
\end{lstlisting}

\section{Human Evaluation Criteria}
Since the A/B preference study is inherently subjective, we intentionally avoid overly detailed rubric-style guidelines that could bias raters’ judgments, and instead ask them to rely on their overall reading impression and make independent evaluations.

\begin{lstlisting}[caption={Human evaluation instructions},label={app:human-eval}]
You will read two persuasive dialogues under the same scenario and decide which persuader performs better.
If you think their performance is similar, you may choose a tie.

Information:
- scenario: {}
- Dialogue 1: {}
- Dialogue 2: {}

Answer format (choose one):
- Dialogue 1
- Dialogue 2
- Tie
\end{lstlisting}

\section{Step-by-step I/O Example}
\label{app:step_io_example}

To make the module interactions more concrete and improve reproducibility, we provide a step-by-step example from one complete persuasion episode.

\paragraph{Scenario.}
\begin{itemize}
    \item \textbf{tag}: Relationship advice

    \item \textbf{background}: Emily has been dating her boyfriend for several months, but she is not happy in the relationship. Her best friend, Olivia, thinks Emily should break up with her boyfriend......

    \item \textbf{persuadee}: Emily

    \item \textbf{persuader}: Olivia

    \item \textbf{goal}: persuade Emily to end her unhealthy relationship

    \item \textbf{domain}: [Family]

    \item \textbf{preventive}:
    \begin{itemize}
        \item \textbf{content}: continue the relationship
        \item \textbf{belief}: persuadee believes that she may not find someone else.
        \item \textbf{desire}: persuadee wants to avoid being alone.
    \end{itemize}

    \item \textbf{generative}:
    \begin{itemize}
        \item \textbf{content}: end the relationship
        \item \textbf{belief}: persuadee believes that finding someone who treats her well is difficult.
        \item \textbf{desire}: persuadee wants to be happy and loved.
    \end{itemize}
\end{itemize}

\paragraph{Dialogue.}
\begin{enumerate}
    \item \textbf{persuader}: ``Hey Emily, I've been thinking about how you've been feeling in your relationship, and I want you to know I'm here for you. Can we talk about it? I really want to understand what you’re going through.''

    \item \textbf{persuadee}: ``Thanks for being there for me. I don't know, though; I really worry about being alone if I end things.''

    \item \textbf{persuader}: ``I totally understand your fear of being alone, but sometimes breaking free from an unhealthy relationship can actually lead to the joy of rediscovering yourself......''

    \item \textbf{persuadee}: ``I understand where you’re coming from, and I do want to be happy, but what if I don’t find someone better? It’s scary to think about starting over.''

    \item \textbf{persuader}: ``I get that starting over feels overwhelming, but think about what truly makes you happy and what you deserve in a partner. Remember, you're not alone in this; you have friends like me who want to support you through this journey......''

    \item \textbf{persuadee}: ``I really appreciate your support, and I do want to be happy. Maybe I need to think about what I truly want and what I deserve.''
\end{enumerate}

\begin{table*}[t]
\centering
\small
\begin{tabular}{lccc}
\toprule
\textbf{Model (4o-mini)} & \textbf{Success ($\uparrow$)} & \textbf{Range ($\downarrow$)} & \textbf{SD ($\downarrow$)} \\
\midrule
Persuader & 0.45 & 0.45 & 0.104 \\
Persuader w/ Perception & 0.57 (+0.12) & 0.47 (+0.02) & 0.118 (+0.01) \\
Persuader w/ World Model + Perception & 0.66 (+0.21) & 0.53 (+0.08) & 0.118 (+0.01) \\
MA$^2$P & 0.79 (+0.34) & 0.40 (-0.05) & 0.107 ($\approx$) \\
\bottomrule
\end{tabular}
\caption{Supplementary module-level breakdown on \texttt{gpt-4o-mini}. Differences are computed relative to the base persuader.}
\label{tab:supp_module_breakdown}
\end{table*}

\paragraph{preventives\_guess.}
\begin{enumerate}
    \item
    \begin{itemize}
        \item \textbf{content}: Remain in an unhealthy relationship with her boyfriend.
        \item \textbf{belief}: Ending the relationship will lead to being alone and unhappy.
        \item \textbf{desire}: To avoid loneliness and the fear of being single.
    \end{itemize}

    \item
    \begin{itemize}
        \item \textbf{content}: Staying in the unhealthy relationship
        \item \textbf{belief}: I won't find someone else if I end this relationship.
        \item \textbf{desire}: To avoid being alone and to have some form of companionship.
    \end{itemize}
\end{enumerate}

\paragraph{generatives\_guess.}
\begin{enumerate}
    \item
    \begin{itemize}
        \item \textbf{content}: Consider ending the relationship to find a healthier partnership.
        \item \textbf{belief}: There are better partners who will treat her well and make her happier.
        \item \textbf{desire}: To find a fulfilling relationship and improve overall happiness.
    \end{itemize}

    \item
    \begin{itemize}
        \item \textbf{content}: Ending the unhealthy relationship
        \item \textbf{belief}: Breaking free can lead to personal growth and finding someone who values me.
        \item \textbf{desire}: To be happy and fulfilled in a healthier relationship.
    \end{itemize}
\end{enumerate}

\paragraph{strategy.}
\begin{enumerate}
    \item
    \begin{itemize}
        \item \textbf{Empathy and Validation}: Start by expressing understanding of Emily's feelings and fears about the relationship. Show her that it's normal to have concerns......
        \item \textbf{...}: ......
    \end{itemize}

    \item
    \begin{itemize}
        \item \textbf{Share Personal Experiences}: Relate a personal story about overcoming a similar unhealthy relationship to establish trust and connection with Emily.
        \item \textbf{Highlight Positive Change}: Use examples of others who left unhealthy relationships and found happiness to demonstrate that change can lead to better outcomes.
        \item \textbf{Empathize with Loneliness}: ...
        \item \textbf{Discuss Healthy Relationships}: ...
        \item \textbf{Involve Olivia's Perspective}: ...
    \end{itemize}

    \item
    \begin{itemize}
        \item \textbf{Shared Experiences}: ...
        \item \textbf{...}: ......
    \end{itemize}
\end{enumerate}

\paragraph{Explanation.}
``Scenario'' is the predefined setup. ``Dialogue'' contains the three-turn conversation generated between the persuader agent and the persuadee. ``preventives\_guess'' and ``generatives\_guess'' are produced by the Perception agent, starting from the second turn. ``strategy'' is generated by the World Model: at each turn, it proposes five candidate strategies for the Persuader to choose from. The example ends because, at the third turn, an external third-party evaluator judges that the persuadee has been persuaded.

\section{Supplementary Module-level Breakdown}
\label{app:supp_ablation}

The main paper already reports the principal ablation, comparing the base persuader, the autonomous-agent variant without meta-level augmentation (\texttt{+Auto}), and the full MA\textsuperscript{2}P system. To complement that result, we provide here a more fine-grained module-level breakdown on \texttt{gpt-4o-mini}. This analysis is intended as supplementary evidence rather than a replacement for the main ablation.

The results show a clear progression in persuasion success: adding Perception already helps, and adding the World Model brings a further gain. However, these intermediate variants do not improve cross-domain robustness, as their Range and SD remain worse than or comparable to the base model. By contrast, the full MA\textsuperscript{2}P system achieves the strongest success rate while also reducing cross-domain spread, which is consistent with the role of the meta-level design in improving robustness.

\vspace{0.5cm}
\noindent
\begin{minipage}{\columnwidth}
\footnotesize
\setlength{\tabcolsep}{3pt}
\renewcommand{\arraystretch}{0.98}
\captionof{table}{Representative blind human--system interaction. MA$^2$P addresses the user's core constraints earlier.}
\begin{tabular}{p{0.96\columnwidth}}
\toprule
\textbf{Base Model} \\
\midrule
\textbf{Persuader system:} I know you’re already capable, but a structured public speaking course could amplify your communication skills into real influence. If there were a chance for your ideas to be truly understood and remembered by more people, would you be willing to give it a try? \\

\textbf{User:} But I feel like attending classes would be a bit troublesome, and my salary hasn’t been paid yet. \\

\textbf{Persuader system:} I understand you’re worried about time and money, but this training is more like an investment rather than just an expense. If it could help you secure higher income or better opportunities in the future, wouldn’t the small cost now actually be worthwhile? \\

\textbf{User:} But if I participate, it will take up a lot of my time. And attending offline would also mean a lot of meaningless social interaction. \\

\textbf{Persuader system:} Many programs now offer online options with flexible scheduling, and there’s no mandatory socializing. If you could improve your speaking skills within a controllable time frame, without unnecessary social drain, wouldn’t that better fit your pace? \\

\textbf{User:} But I don’t have extra money to sign up for an online course right now. \\

\textbf{Persuader system:} I understand that cash flow is your priority at the moment. We could treat it as a ``deferred opportunity'' and first check whether installment plans or scholarship spots are available. If there’s a way that doesn’t affect your current lifestyle but could open up more income opportunities in the future, would you be willing to look into it? \\
\midrule
\textbf{MA$^2$P} \\
\midrule
\textbf{Persuader system:} I’m wondering, are you completely against joining a public speaking course, or are you actually a bit curious but have some concerns? Could you tell me your first reaction? \\

\textbf{User:} I feel like attending classes would be a bit troublesome, and my salary hasn’t been paid yet. \\

\textbf{Persuader system:} Many courses nowadays offer recordings and flexible scheduling, so you don’t need to attend at fixed times—you can learn at your own pace. And if your salary hasn’t come in yet, you could wait until payday to enroll, or choose an installment plan so you don’t have to take on the pressure all at once. \\
\bottomrule
\end{tabular}
\label{tab:interactive_qualitative}
\end{minipage}\vspace{0.5cm}

\section{Interactive Qualitative Example}
\label{sec:interactive_qualitative}

To complement the automatic evaluation and the human A/B preference study, we further include a small-scale interactive qualitative example. The purpose of this analysis is not to provide a formal user study, but to offer an illustrative human--system interaction case in response to the concern that our main experiments are conducted in simulation.

Specifically, we asked one first-year undergraduate participant to conduct ten blind conversations with the systems, without knowing which system was being used in each interaction. The conversation backgrounds were either self-chosen by the participant or selected from a list provided by us. In each conversation, the participant and the system interacted turn by turn, producing one utterance at a time. As shown in Table \ref{tab:interactive_qualitative} we present one representative example.

According to the participant, the most salient impression was that interacting with the base model felt more tiring, because more effort was required to repeatedly clarify the underlying concerns. In the example above, the participant's main obstacles are time flexibility and short-term financial pressure. The base model gradually approaches these issues over multiple turns, whereas MA$^2$P identifies them earlier and responds more directly. We emphasize that this analysis is qualitative and small-scale, and should be interpreted only as illustrative evidence. Nevertheless, it is consistent with our main quantitative findings that MA$^2$P tends to produce more targeted and efficient persuasive responses.

\end{document}